\documentclass[format=acmsmall, review=false, screen=true]{acmart}

\usepackage{booktabs} % For formal tables

\usepackage[linesnumbered,boxed,ruled,vlined]{algorithm2e}

\usepackage{amsmath,bm}

\usepackage{array}

\usepackage{graphicx, subfigure}

\usepackage{url,soul}

\usepackage{enumitem}

\usepackage{multirow,array}

\usepackage{mathrsfs}

\usepackage{hyperref}

\usepackage[perpage,symbol]{footmisc}
\setfnsymbol{wiley}
\usepackage{dsfont,amssymb}

\newcommand{\stitle}[1]{\vspace{0.5mm} \noindent {\bf{#1}} }
\newcommand{\sstitle}[1]{\vspace{0.5mm} \noindent {\bf\emph{#1}}}

\setcopyright{acmcopyright}
\acmJournal{TKDD}
\acmYear{2018} \acmVolume{1} \acmNumber{1} \acmArticle{1} \acmMonth{1} \acmPrice{15.00}\acmDOI{10.1145/3264745}
\received{July 2017}
\received[revised]{Apr 2018}
\received[accepted]{Aug 2018}

\begin{document}
% Title portion. Note the short title for running heads 
\title[Semi-supervised Learning Meets Factorization]{Semi-supervised Learning Meets Factorization: Learning to Recommend with Chain Graph Model}

\author{Chaochao Chen}
\affiliation{%
  \institution{Zhejiang University}
  \city{Hangzhou}
  \country{China}
  }
  \affiliation{%
  \institution{University of Illinois at Urbana-Champaign}
  \city{Urbana}
  \state{IL}
  \postcode{61802}
  \country{USA}
  }
\author{Kevin Chen-Chuan Chang}
\affiliation{%
  \institution{University of Illinois at Urbana-Champaign}
  \city{Urbana}
  \state{IL}
  \postcode{61802}
  \country{USA}
}
\author{Qibing Li}
\author{Xiaolin Zheng*}\footnote{Corresponding author. }
\affiliation{%
  \institution{College of Computer Science, Zhejiang University}
  \city{Hangzhou}
  \country{China}
}

\begin{abstract}
Recently \emph{latent factor model} (LFM) has been drawing much attention in recommender systems due to its good performance and scalability.
However, existing LFMs predict missing values in a user-item rating matrix only based on the known ones, and thus the \emph{sparsity} of the rating matrix always limits their performance.
Meanwhile, \emph{semi-supervised learning} (SSL) provides an effective way to alleviate the label (i.e., rating) sparsity problem by performing label propagation, which is mainly based on the \emph{smoothness} insight on affinity graphs.
However, graph-based SSL suffers serious scalability and graph unreliable problems when directly being applied to do recommendation.
In this paper, we propose a novel probabilistic \emph{chain graph model} (CGM) to marry SSL with LFM. The proposed CGM is a combination of \emph{Bayesian network} and \emph{Markov random field}. The Bayesian network is used to model the rating generation and regression procedures, and the Markov random field is used to model the confidence-aware smoothness constraint between the generated ratings. 
Experimental results show that our proposed CGM significantly outperforms the state-of-the-art approaches in terms of four evaluation metrics, and with a larger performance margin when data sparsity increases.
\end{abstract}

%
% The code below should be generated by the tool at
% http://dl.acm.org/ccs.cfm
% Please copy and paste the code instead of the example below. 
%


%
% End generated code
%


\thanks{This work was supported in part by the National Natural Science Foundation of China (No.U1509221), the National Key Technology R\&D Program (2015BAH07F01), and the Zhejiang Province key R\&D program (No.2017C03044).}

\maketitle

% The default list of authors is too long for headers}
%\renewcommand{\shortauthors}{C.C. Chen et al.}

\section{Introduction}\label{intro}
As e-commerce platforms and online social networks provide customers and users myriad products and massive information, a \emph{recommender system} (RS) has become a useful and even indispensable facility for information filtering.
In fact, RS is now part of our daily activities, as it has been widely adopted by internet service leaders, such as Amazon, Youtube, and Facebook \cite{sarwar2001item}.

\stitle{Latent factor model.} 
Among the existing recommendation approaches, latent factor model (LFM) has been drawing much attention due to its good performance and scalability. 
LFM uses a low dimensional user and item latent factors to represent the characteristics of each user and each item, and uses the product of them to represent the user's rating on the item. LFM has drawn much attention recently due to its good performance and scalability \cite{mnih2007probabilistic,koren2008factorization,takacs2008investigation,agarwal2009regression,rendle2010factorization,wang2011collaborative,purushotham2012collaborative,kanagal2013focused,xu2015ice,liu2016kernelized,guo2016novel,beutel2017beyond}. 
As \emph{users} have actions (e.g., rate and buy) on some \emph{items}, LFM aims to predict the users' unknown actions on other items.
The tendency of a user's action on an item can be indicated by a real-valued number, i.e., \emph{rating} or \emph{label}. Thus, the recommendation problem is also known as the \emph{unknown ratings prediction} problem \cite{shi2014collaborative}. 
In practice, however, many LFMs have to evaluate very large user and item sets, where the user-item (U-I) matrix is extremely sparse-- such \emph{data sparsity} has always been its main challenge \cite{su2009survey}.

%Given a user set $\mathds{U}=\{u_1,...,u_I\}$ and an item set $\mathds{V}=\{v_1,...,v_J\}$, where \emph{I} and \emph{J} denote the number of users and items.
%We use $\bm{R}$ to represent the user-item (U-I) rating matrix, with each element $R_{ij}$ denoting the rating that $u_i$ gives to $v_j$. In practice, $R_{ij}$ is often given as integers, such as in the range [1, 5] on MovieLens. We use $\bm{r}$ to represent the predicted U-I rating matrix, with a real-valued number $r_{ij}$ denoting the predicted rating of $u_i$ on $v_j$. There are totally $I \times J$ ratings, among which only $\mathscr{L}$ ratings are known, and our recommendation task is to predict the unknown ratings in $\bm{R}$, with the size of $\mathscr{U}=I \times J-\mathscr{L}$.

%\stitle{Data sparsity.}

%So far, two techniques have been popularly used to alleviate the data sparsity problem of RS, i.e., \emph{semi-supervised learning} (SSL) and \emph{latent factor model} (LFM).

\stitle{Semi-supervised learning.} 
SSL uses unlabeled data to either modify or reprioritize hypotheses obtained from labeled data alone, and thus can alleviate the label sparsity problem by adopting the graph information between data \cite{seeger2000learning}. 
%Note that `rating' is called from the perspective of RS and `label' is called from that of SSL, and we use them interchangeably in this paper. 
Towards effective SSL, affinity graph-based smoothness approaches have attracted much research interests, which follow the \emph{smoothness} insight: \emph{close nodes on an affinity graph have similar labels}. 
%The affinity graph is used to capture the affinity between data. 
Graph-based SSL is appealing recently because it is easy to implement and gives rise to closed-form solutions \cite{zhu2003semi,chapelle2006semi,wang2006recommendation,ding2007learning,wang2010enhanced,fang2014graph}. 
However, graph-based SSL directly predicts the unknown ratings in the original U-I matrix, and thus suffers from the scalability problem. 
%In a RS scenario, each data object $(u_i,v_j)$ is a U-I pair, with a rating $R_{ij}$ as its label, and thus, to adopt SSL for rating prediction in RS, rating smoothness should exist among U-I pairs.

%LFM increases the rating density by reducing the dimensionality of U-I rating matrix, and thus is able to alleviate the data sparsity problem.  

%Although SSL and LFM have their own merits to alleviate the data sparsity problem, they  have their own disadvantages. 
%Graph-based SSL directly predicts the unknown ratings in the original U-I matrix, and thus suffers from the scalability problem. Meanwhile, LFM focuses on regressing the rare labelled data, and fails to capture the smoothness insight between ratings, unlike SSL, and thus its performance is further limited under data sparsity scenario \cite{agarwal2009regression}.

%LFM scales linearly with the labelled data size due to the use of dimensionality reduction technique \cite{koren2008factorization,agarwal2009regression,rendle2010factorization}, and thus can solve the scalability problem of SSL. 
%Because LFM predicts ratings by learning low dimensional user and item latent factors,  instead of predicting the ratings directly, which is the same reason as it solves the scalability problem of memory-based collaborative filtering approach \cite{su2009survey}.

As a key insight of this paper, we identify the marriage of SSL and LFM. 
The main insights of SSL (i.e., \emph{smoothness}) and LFM (i.e., \emph{dimensionality reduction}) are orthogonal, and they are likely to benefit from each other. However, surprisingly, the synergy between SSL and LFM has not been explored. 
On one hand, the smoothness idea of SSL, i.e., similar users or items should give or receive similar ratings-- across the entire U-I matrix of ratings, known or unknown, can mitigate the data sparsity problem of LFM. On the other hand, the dimensionality reduction idea of LFM can solve the scalability problem of SSL, since the prediction is done in a low rank matrix instead of the original high dimensional matrix. 
%We will analyze our model time complexity in Section \hyperref[parameter-learning]{\ref{parameter-learning}}.

\stitle{Challenges of marrying SSL with LFM.}However, marrying SSL with LFM to do recommendation is nontrivial. 
We summarize the main challenges as follows:
%Recently, there is an attempt of using the dimensionality reduction idea of LFM to realize the similar idea of SSL \cite{ma2011recommender,gu2010collaborative,chen2014context}. They assume that connected users or items on a graph tend to share similar user or item latent factors, which we term it ``\emph{latent factor restriction}'' (LFR) hereafter. That is, they capture the correlation dependency between latent factors, not ratings. We will explain in Section \hyperref[relatedwork-lfr]{\ref{relatedwork-lfr}} that they use an overly strong assumption which limits their performance, and experiments in Section \hyperref[experiments]{\ref{experiments}} will demonstrate their deficiency, especially in data sparsity scenarios. Moreover, several efforts have attempted to directly adopt SSL to do recommendation without using the dimensionality reduction idea of LFM  \cite{wang2006recommendation,ding2007learning,wang2010enhanced}. They first build a user or an item social network based on the rare user social relationships or ratings, and then perform rating propagation on them. They not only overlook that the networks are built based on the rare information and thus are unreliable, but also neglect the reality that the rating smoothness in a RS scenario should be between U-I pairs, instead of only users or items, as we described above.

\sstitle{Challenge $\mathcal{I}$: Disparate unification.}~SSL predicts unknown ratings through rating propagation on affinity graphs. I.e., SSL captures the correlation dependency between ratings. LFM predicts unknown ratings through learning user and item latent factors by regressing known ratings. I.e., LFM captures the causal dependency between latent factors and ratings. 
%Usually, a single model only has the ability to capture single dependency between variables, e.g., \cite{mnih2007probabilistic,koren2008factorization,agarwal2009regression,rendle2010factorization,wang2011collaborative}.
%Thus, how to build a unified model so that such different dependency between different variables can be captured is the first challenge.
Therefore, we aim to propose a principled framework to unify SSL and LFM so that such different dependency between different variables can be captured.

\sstitle{Challenge $\mathcal{II}$: Expensive graph construction.}~In a RS scenario, each data object $(u_i,v_j)$ is a U-I pair, with a rating $R_{ij}$ as its label, and thus, to adopt SSL for rating prediction in RS, rating smoothness should exist among U-I pairs.
%In a RS scenario, each data object is a U-I pair, e.g., $\{u_i,v_j\}$, with a rating as its label, e.g., $R_{ij}$, and thus, to adopt SSL for rating prediction in RS, we need to perform rating smoothness among U-I pairs. 
Suppose we have \emph{I} users and \emph{J} items, building such a U-I pairwise affinity graph needs to compute the similarity between each of $ I \times J$ U-I pairs, with a time complexity of $O{(I \times J)}^2$, and thus, cannot scale to large datasets. 
%Moreover, as described above, directly using graph-based SSL to do prediction needs a cubic time complexity $O(\mathscr{U}^3)$ where $\mathscr{U}$ is a very large.
%Thus how to find a smoothness approach which possesses low time complexity and similar performance with pairwise smoothness becomes the second challenge.
Thus, we aim to realize smoothness in a more efficient manner.

\sstitle{Challenge $\mathcal{III}$: Unreliable affinity.}~In a traditional SSL scenario, the affinity graphs are built based on the characteristics of the data itself, e.g., the pixel data of a scanned digit. In contrast, in a RS scenario, the affinity graphs are usually built based on user social relationships or ratings \cite{wang2006recommendation,ding2007learning,wang2010enhanced,purushotham2012collaborative,ma2011recommender,gu2010collaborative,chen2014context}. As a result, the affinity graphs are unreliable due to the sparsity of such user social relationships or ratings (U-I pairs). 
Thus, we aim to alleviate the unreliable affinity problem in a robust way.
%How to perform rating propagation on an unreliable affinity graph is the third great challenge. 

\stitle{Our proposal: Learning to recommend with CGM.}Our proposal is based on the following insights.
%In this paper, we propose a novel chain graph model (CGM), which is a combination of Bayesian network and Markov random field, and is able to marry SSL and LFM successfully.
%which are based on the following insights:

\sstitle{Insight $\mathcal{I}$: Principled unification.}~To address Challenge $\mathcal{I}$, we propose a novel chain graph model (CGM) to marry SSL with LFM. As far as we know, this is the first attempt in the literature to adopt CGM in RS. A CGM is a combination of Bayesian network and Markov random field, which has the ability to capture different kinds of dependency (i.e., correlation and causal dependency) between different kinds of variables (i.e., latent factors and ratings) \cite{lauritzen2002chain,bishop2006pattern,mackey2007latent}. Thus, CGM is an ideal model to solve Challenge $\mathcal{I}$.

%we need a way that not only can capture different kinds of dependency (i.e., correlation dependency and causal dependency), but also is able to capture the dependency between different kinds of variables (i.e., latent factor and rating). 
%Bayesian network is commonly used to model the rating generation and regression from user and item latent factors \cite{mnih2007probabilistic,wang2011collaborative,purushotham2012collaborative}. Markov random field is commonly used to model the correlation dependency between variables \cite{domingos2001mining}. 
%As a combination of Bayesian network and Markov random field, 

\sstitle{Insight $\mathcal{II}$: Efficient smoothness.}~To address Challenge $\mathcal{II}$, we develop a novel ``joint smoothness'' framework to realize SSL on a pair of decomposed user and item affinity graphs.
%In the RS scenario, each data object is a U-I pair, and this results in Challenge $\mathcal{II}$. Although pairwise smoothness is a standard way to perform smoothness, it results in high time complexity. 
Since a rating is given from a user to an item, smoothness exists in two dimensions, i.e., user and item. We term it \emph{joint-smoothness}, which enables us to decrease the time complexity to build affinitive graphs to $O(I^2 + J^2)$. 
%Thus, joint-smoothness is a way to solve Challenge $\mathcal{II}$.

\sstitle{Insight $\mathcal{III}$: Selective smoothness.}~To address Challenge $\mathcal{III}$, we propose a confidence-aware smoothness approach. 
%To our knowledge, our effort is the first in the literature to identify the unreliable affinity  problem in RS and propose a solution.
Different from the traditional SSL which performs rating propagation (i.e., smoothness constraint) everywhere on the affinity graphs, we choose to perform smoothness selectively. 
Specifically, we propose a smoothness confidence decay model to control the hops of rating propagation length.%enables us to only propagate the known ratings to their close neighbors.
%Thus, a confidence-aware selective smoothness method is an effective way to solve Challenge $\mathcal{III}$.

\sstitle{Solutions.}~First, we propose a novel CGM (Section \hyperref[cfssl]{\ref{cfssl}}), which is a combination of Bayesian network and Markov random field.
%Our proposed CGM contains three layers. The first layer is the \emph{latent factor layer}. The second layer is the \emph{prediction layer}. The third layer is the \emph{observation layer}. From the latent factor layer to the prediction layer, it is the rating generation procedure. In the prediction layer, it is the prediction smoothness constraint of SSL. From the prediction layer to the observation layer, it is the rating regression objective of the existing LFMs. %, i.e., the predicted rating should be close to the observed rating. 
%We use Bayesian network to model the rating generation and regression procedures, and use Markov random field to model the rating smoothness constrain. 
Second, we propose a joint-smoothness objective function. Instead of building a U-I pairwise affinity graph, we build two decomposed user and item affinity graphs. 
%We first describe the derivation from pairwise to joint smoothness (Section \hyperref[pairwise-to-joint]{\ref{pairwise-to-joint}}), 
%and then present the affinity graphs build method (Section \hyperref[sec-energyfunction-buildgraph]{\ref{sec-energyfunction-buildgraph}}).
Third, we propose a confidence-aware smoothness approach (Section \hyperref[confident-pairwise-to-joint]{\ref{confident-pairwise-to-joint}}). This selective smoothness approach not only alleviates the graph unreliable problem in RS scenarios, but also saves huge computation. 
%The dimensionality idea of LFM and the confidence-aware smoothness idea will make our model scales linear with the observed data size.
Finally, we present the model learning method based on coordinate descent (Section \hyperref[parameter-learning]{\ref{parameter-learning}}).

%Specifically, we build two undirected user and item decomposed affinity graphs, i.e., Markov random fields, to represent the affinitive relationship between users and items. Finally, the posterior distribution over the user and item latent factors can be obtained through Bayesian theory, and then it can be solved by using coordinate descent. Getting user and item latent factors will enable us to predict any unknown U-I rating pairs.

%we propose a confidence-based joint-smoothness energy function to model the probability of smoothness relationship between the generated U-I ratings pairs on user and item affinity graphs (Section V). That is, we use Markov random field to model the smoothness constrain between the generated ratings on the user and item affinity graphs. However, different from the traditional SSL that performs rating propagation (i.e., smoothness constrain) everywhere on the affinity graphs, we only propagate the known ratings to their close neighbors, where we have strong confidence. By doing this, we are able to not only alleviate the graph unreliable problem in RS scenario, but also save huge computation.

%Our proposed chain graph model realizes the idea of both SSL and LFM, and thus possesses the merits of both of them. On one hand, the data sparsity issue of LFMs is alleviated by using the smoothness idea of SSL. One the other hand, the scalability problem of SSL is solved by using the dimensionality reduction technique of LFMs.

\sstitle{Results.}~We concretely realize our solutions of marrying SSL with two kinds of popular LFMs, and conduct comprehensive experiments in Section \hyperref[experiments]{\ref{experiments}}. Our experiments are conducted on \emph{three} popular datasets, with \emph{nine} state-of-the-art comparison approaches. We use \emph{four} metrics to evaluate model performance. The experimental results show that our approach significantly outperforms the state-of-the-art methods, especially in data sparsity scenarios.

\stitle{Contributions.}We summarize the main contributions in this paper as follows: 
\begin{itemize}[leftmargin=*] \setlength{\itemsep}{-\itemsep}
    \item We propose a novel CGM to marry SSL with LFM for alleviating the data sparsity problem of RS, which we believe is the first attempt in the literature.
    \item We propose to perform joint-smoothness instead of pairwise smoothness, which has better efficiency.
    \item We propose a confidence-aware smoothness approach to alleviate the unreliable graph problem in RS scenario. To the best of our knowledge, it is also the first attempt.
   \item Our model scales linearly with the observed data size, since we adopt dimensionality reduction technique and confidence-aware smoothness approach.
\end{itemize}

%(1) We propose a novel CGM to marry SSL and LFM for alleviating the data sparsity problem of RS, and we also present the model learning methods using coordinate descent (Section \hyperref[parameter-learning]{\ref{parameter-learning}}). As far as we know, this is the first attempt in the literature to adopt CGM in RS; (2) We propose to perform joint-smoothness instead of pairwise smoothness; (3) We propose a confidence-aware joint-smoothness energy function to alleviate the graph unreliable problem in RS scenario. To the best of our knowledge, it is the first attempt in the literature to identify the affinity unreliable problem in RS and propose an approach to solve it; (4) Our models can be applied to very large datasets since it scales linearly with the total number of observed data.

The rest of the paper is organized as follows. In Section \hyperref[relatedwork]{\ref{relatedwork}}, we review related work. 
In Section \hyperref[preliminary]{\ref{preliminary}}, we describe the popular realizations of SSL and LFM. 
In Section \hyperref[cfssl]{\ref{cfssl}}, we propose a novel probabilistic CGM to marry SSL with LFM.
In Section \hyperref[sec-energyfunction]{\ref{sec-energyfunction}}, we present the confidence-aware joint-smoothness energy function.
In Section \hyperref[parameter-learning]{\ref{parameter-learning}}, we present the model learning approach based on gradient descent. 
In Section \hyperref[experiments]{\ref{experiments}}, we present the experimental results and analysis. 
Finally, we conclude the paper in Section \hyperref[conclusion]{\ref{conclusion}}.

\section{Related Work}\label{relatedwork}

As described in Section \hyperref[intro]{\ref{intro}}, %semi-supervised learning (SSL) and latent factor model (LFM) both have their advantages and shortcomings when making rating prediction. LFR attempts to adopt the idea of SSL in LFM, but it can not deal with Challenges $\mathcal{I}$, $\mathcal{II}$, and $\mathcal{III}$. 
we want to marry SSL with LFM by using chain graph model (CGM), in this section, we review literature on SSL, LFM, LFR, and CGM, respectively.

\subsection{Semi-supervised learning}\label{relatedwork-ssl}
SSL uses unlabeled data to either modify or reprioritize hypotheses obtained from labeled data alone \cite{seeger2000learning}. 
Among the existing SSL techniques, graph-based SSL is the most promising one, which alleviates data sparsity problem by performing label propagation on the affinity graphs, and its main insight is graph-based smoothness \cite{zhu2003semi,fang2014graph}. 
%RS and SSL have a lot in common, e.g., they both can be seen as a label prediction problem. 
In the literature, several different objective functions are proposed to realize graph-based smoothness, e.g., Harmonic Function (HF) \cite{zhu2003semi} and Green's Function (GF) \cite{ding2007learning}. 
Take HF for example, it minimizes the rating difference between close nodes and thus achieves smoothness on $\mathcal{G}$, which is the same as propagate the known labels to the unknown ones on the affinity graph \cite{zhu2002learning}.
So far, there are several research directly adopting SSL to do rating prediction \cite{wang2006recommendation,ding2007learning,wang2010enhanced}. However, as described in Section \hyperref[intro]{\ref{intro}}, directly adopting them in RS suffers from scalability and unreliable affinity problems. 
%Since we are trying to adopt the smoothness idea of SSL in this paper, we will further describe how to realize SSL smoothness in a RS scenario in Section \hyperref[preliminary-ssl]{\ref{preliminary-ssl}}.

\subsection{Latent factor model.}

Generally, existing LFMs can be divided into three types: basic matrix factorization (MF) that only uses U-I rating matrix to do prediction, side information aided MF that uses other side information besides the U-I rating matrix, and latent factor restriction models that consider user/item affinity graphs. 

\sstitle{Basic matrix factorization.}~MF has drawn much attention recently since it was adopted in the Netflix competition \cite{koren2008factorization}, and its main insight is the dimensionality reduction technique \cite{sarwar2000application}. The most basic MF model, known as probabilistic matrix factorization (PMF) or single value decomposition (SVD), factorizes a U-I matrix into a low rank user feature matrix and item feature matrix, and then uses their product to predict unknown ratings \cite{mnih2007probabilistic,takacs2008investigation,xu2015ice,hernando2016non,liu2016kernelized,beutel2017beyond,feng2017recommendations}.
Other promotions of SVD include SVDB and SVD++, which further consider user/item rating bias and implicit feedback information \cite{koren2008factorization}.

\sstitle{Matrix factorization with side information.}~Generally, side information can be divided into three categories: content information, social information, and other context information, e.g., user attributes. These three kinds of information have been proven efficient to improve recommendation performance \cite{agarwal2010flda,wang2011collaborative,mcauley2013hidden,chen2014context,agarwal2009regression,rendle2010factorization,agarwal2010flda,jamali2010matrix,jiang2014scalable,guo2016novel,chen2016capturing,chen2016recommender,yan2017approach}. 

First, regression-based latent factor model (RLFM) \cite{agarwal2009regression} and factorization machines \cite{rendle2010factorization} were proposed to incorporate context information to improve recommendation performance. However, these context information are not always available and hard to obtain, and thus they are out of the scope of this paper. 
Second, the integration of factor model and LDA (fLDA) \cite{agarwal2010flda} further incorporates item content information into RLFM. 
Third, Wang et al. propose collaborative topic regression (CTR) \cite{wang2011collaborative}, which systematically combines PMF and latent Dirichlet allocation (LDA) \cite{blei2003latent}. CTR uses PMF to factorize U-I rating information and uses LDA to mine item content information. It has been proven that CTR outperforms fLDA in a similar setting, since fLDA largely ignores the other users ratings \cite{wang2011collaborative}. 
%In this paper, we take CTR as a special case of LFM and will further describe it in details in Section \hyperref[preliminary-ctr]{III}. 
Later, CTR-SMF \cite{purushotham2012collaborative} was proposed to factorize not only rating information but also user social information to make a better recommendation.

However, existing LFMs only fit the model by minimizing the difference between the rare known ratings and the predicted ratings, and do not consider rating smoothness nature between similar U-I pairs.

\sstitle{Latent factor restriction.}\label{relatedwork-lfr} 
The most similar works to ours are LFRs \cite{ma2011recommender,gu2010collaborative,chen2014context,rao2015collaborative}.
They first use rating or side information to build user or item affinity graphs, and then constrain latent factor smoothness on the graphs. They assume that connected user or item on the affinity graphs should have similar latent factors. We divide the existing LFRs into two types, i.e, user latent factor restriction (ULFR) \cite{ma2011recommender,chen2014context} and user-item latent factor restriction (UILFR) \cite{gu2010collaborative}, which means that only user latent factor and both user and item latent factors are restricted, respectively.

LFR is actually \emph{not} the smoothness insight as captured in SSL, i.e., rating smoothness.
LFR is much stronger than rating smoothness constrain, because LFR leads to rating smoothness, but not the other way around. Take an item affinity graph for example: based on the assumption of LFR, all the ratings of the connected items should be similar, since neighbors have similar latent factors. That is, LFR indicates smoothness exists everywhere on an affinity graph.
Our experiments in Section \hyperref[experiments]{\ref{experiments}} will demonstrate that this overly strong assumption will fail particularly in data sparsity scenarios. 

\subsection{Chain Graph Models.}\label{relatedwork-cgm} 
CGM is a probabilistic model that combines \emph{Bayesian network} and \emph{Markov random field}. Bayesian network is useful to express causal relationships between random variables \cite{bishop2006pattern}, and it is popularly used in the existing LFMs \cite{mnih2007probabilistic,agarwal2009regression,wang2011collaborative}, which use Bayesian network to express the rating generation procedure. Markov random field is suited to express soft constraints between random variables \cite{bishop2006pattern}, and its applications also include recommender systems \cite{domingos2001mining}. Chain graph models contain both Bayesian network and Markov random field, and thus can represent a broader class of distributions, and it has been used in applications include image de-noising \cite{mackey2007latent}, while, surprisingly, not in RS so far.

There are existing works try to bridge collaborative filtering and SSL \cite{cao2013shilling,zhang2014addressing,yang2017bridging}. 
For example, Semi-SAD \cite{cao2013shilling} tried to combine the idea of collaborative filtering and SSL for shilling attack detection tasks. 
The semi-supervised co-training (CSEL) framework \cite{zhang2014addressing} first proposes two context-aware factorization models by leveraging “more general” sources such as age and gender of a user, or the genres. Then, it builds a semi-supervised learning process by assembling two models generated with the above context-aware model. This method belongs to context-aware recommendation approaches and needs to use contextual information, e.g., user age and item category, which is out-of-the-scope of our paper. 
Preference and context embedding (PACE) \cite{yang2017bridging} aim to bridge collaborative filtering and graph-based SSL to do point-of-interest (POI) recommendation through a neural approach, which is the most relevant and state-of-the art model. 
Specifically, PACE uses Skipgram \cite{mikolov2013distributed} to model graph-based context information as the SSL part, and uses a multi-layer neural network to model user-POI interaction information as the collaborative filtering part. 

In this paper, we propose a novel CGM to marry LFM with SSL to do the geneal recommendation tasks, and we also propose a confidence-aware approach to solve the overly strong assumption of LFR. 
Our work is different from it mainly from the following three aspects: 
(1) PACE only aims to do POI recommendation, while our model can be applied into most recommendation scenarios; 
(2) The objective function of PACE is designed for binary ratings (i.e., a rating $r_{ij} \in \{0, 1\}$), while our model is suitable for any rating scales (i.e., a rating $r_{ij} \in \mathds{R}$); 
(3) PACE assumes the affinity graph is reliable due to its application to POI recommendation where affinity graph can be built using location information. 
In contrast, as one of the main contributions of our work, we propose a solution for alleviating the graph unreliable problem in many general recommendation situations. 
We will show the comparison results in experiments: the performance of PACE is limited in the general recommendation scenarios where affinity graph is unreliable.

\section{Preliminary}\label{preliminary}
Since both SSL and LFM can be used to alleviate the data sparsity problem of RS, in this section, we will present the popular approaches of both of them.

\stitle{Problem setting.}In RS, each data object, i.e., a U-I pair, is related to two elements, i.e., user and item. The data label, i.e., U-I rating, is generated from a user to an item. 
Suppose there is a U-I \emph{pairwise affinity graph} $\mathcal{G}=(\mathcal{V}, \mathcal{E})$ with nodes $\mathcal{V}$ denote data objects, edges $\mathcal{E}$ denote the affinitive relationship between nodes, and weights of edges $P$ denote the affinitive relation strength between nodes. 
Let $\bm{U}\in{}\mathds{R}^{K\times{}I}$ and $\bm{V}\in{}\mathds{R}^{K\times{}J}$  be the user and item latent feature matrixes, with their column vectors $\bm{U_i}$ and $V_j$ representing the \emph{K}-dimensional latent vectors of $u_i$ and $v_j$ respectively.
%, and the elements of $U_i$ measure the extent of interests $u_i$ has in the corresponding elements of items.
%, positive or negative.
%Let $\bm{V}\in{}\mathds{R}^{K\times{}J}$ be an item latent feature matrix, where the column vector $\bm{V_j}$ represents the \emph{K}-dimensional item latent vector of $v_j$.
%, and the elements of $V_j$ measure the extent to which $V_j$ possesses those latent characteristics.
%, again positive or negative.

\stitle{Semi-supervised learning approaches.}\label{preliminary-ssl}Graph-based SSL mainly alleviates the data sparsity problem by realizing the smoothness insight on affinity graphs, i.e., close nodes on an affinity graph should have similar labels \cite{zhu2003semi,fang2014graph}. 
In the literature, several different objective functions are proposed to realize graph-based smoothness, e.g., Harmonic Function (HF) \cite{zhu2003semi} and Green's Function (GF) \cite{ding2007learning}. Take HF for example, its energy function on a pairwise affinity graph $\mathcal{G}$ is
\begin{equation}\label{ssl-pair}\small
{{\mathcal L}_P} = \frac{\lambda_P}{2}\sum\limits_{i = 1}^I \sum\limits_{k = 1}^I \sum\limits_{j = 1}^J \sum\limits_{o = 1}^J {P_{ij,ko}}{{\left( {r_{ij} - r_{ko}} \right)}^2},
\end{equation}
where $\{i,j\}$ and $\{k,o\}$ are two nodes on $\mathcal{G}$ and $P_{ij,ko}$ is the weight between them. $\lambda_P$ controls the global smoothness degree on $\mathcal{G}$ and a bigger $\lambda_P$ corresponding to a higher rating smoothness degree on $\mathcal{G}$. Eq.~(\ref{ssl-pair}) minimize the rating difference between close nodes and thus achieves smoothness on $\mathcal{G}$, which is the same as propagate the known labels to the unknown ones on the affinity graph \cite{zhu2002learning}.

\stitle{Latent factor models.}\label{preliminary-ctr}As we described in Section \hyperref[relatedwork]{\ref{relatedwork}}, two promising kinds of LFMs are the basic MF and side information-aided MF. 
The basic MF approach uses $r_{ij} =\bm{{U_i}}^T\bm{V_j}$ to capture $u_i$'s overall interests in $v_j$'s characteristics, that is the predicted rating of $u_i$ on $v_j$ \cite{mnih2007probabilistic}. 
Its object is to minimize the difference between the observed ratings and the predicted ratings:
\begin{equation}\label{pmf}\small
\mathop {\arg \min }\limits_{\bm{U},\bm{V}} \sum\limits_{i = 1}^I {\sum\limits_{j = 1}^J {{I_{ij}^R}{{\left( {{R_{ij}} - r_{ij}} \right)}^2}} }  + {\lambda _U}\sum\limits_{i = 1}^I {||{\bm{U_i}}||_F^2}  + {\lambda _V}\sum\limits_{j = 1}^J {||{\bm{V_j}}||_F^2},
\end{equation}
where $I_{ij}^R$ is an indicator function that equals to 1 if $u_i$ rated $v_j$, 0 otherwise, $||\cdot||_F^2$ denotes the Frobenius norm, and $\lambda_U$ and $\lambda_V$ are regularization
parameters to avoid overfitting.

The side information-aided MF incorporate other information, e.g., item content and context information, to learn a better prior for user and item latent factors. For example, CTR \cite{wang2011collaborative} additionally includes a topic proportion which learned from item's content using topic modeling when modeling item latent factor. 
For its model details, please refer to \cite{wang2011collaborative}.

From the above explanation of the existing LFMs, we can see that they only focus on fitting the model by regressing the rare known ratings, and neglect the rating smoothness insight between similar users and items.

\section{Proposed Chain Graph Model}\label{cfssl}

%\begin{figure}
%\centering
%\includegraphics[width=4.5cm]{figures/model}
%\vskip -0.05in
%\caption{Example of the proposed CGM.}
%\label{cgm-ctr}
%\vskip -0.1in
%\end{figure}

\begin{figure*}
\centering
\subfigure[U-I rating matrix] { \includegraphics[width=3cm]{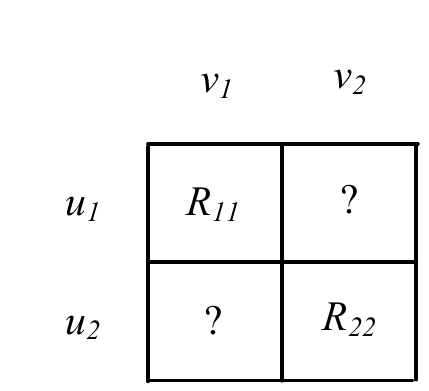}}
~~~~~
\subfigure[SSL example] { \includegraphics[width=3.5cm]{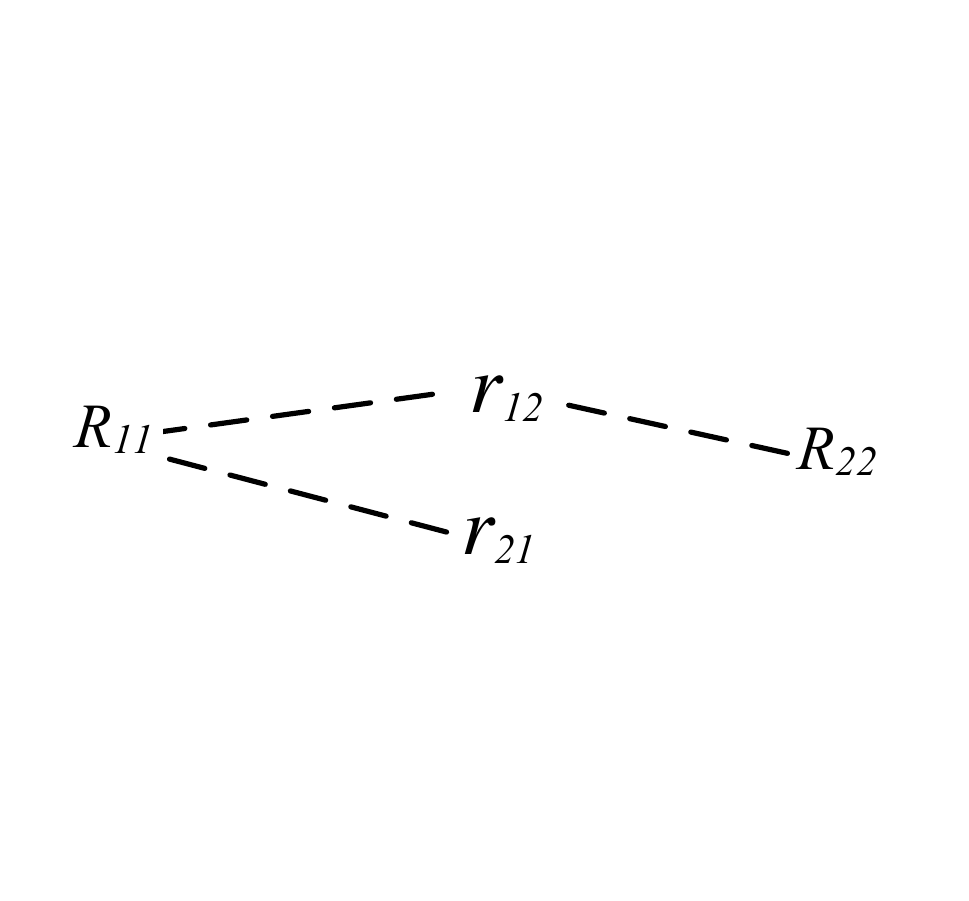}}
~~~~~
\subfigure[LFM example] { \includegraphics[width=3.5cm]{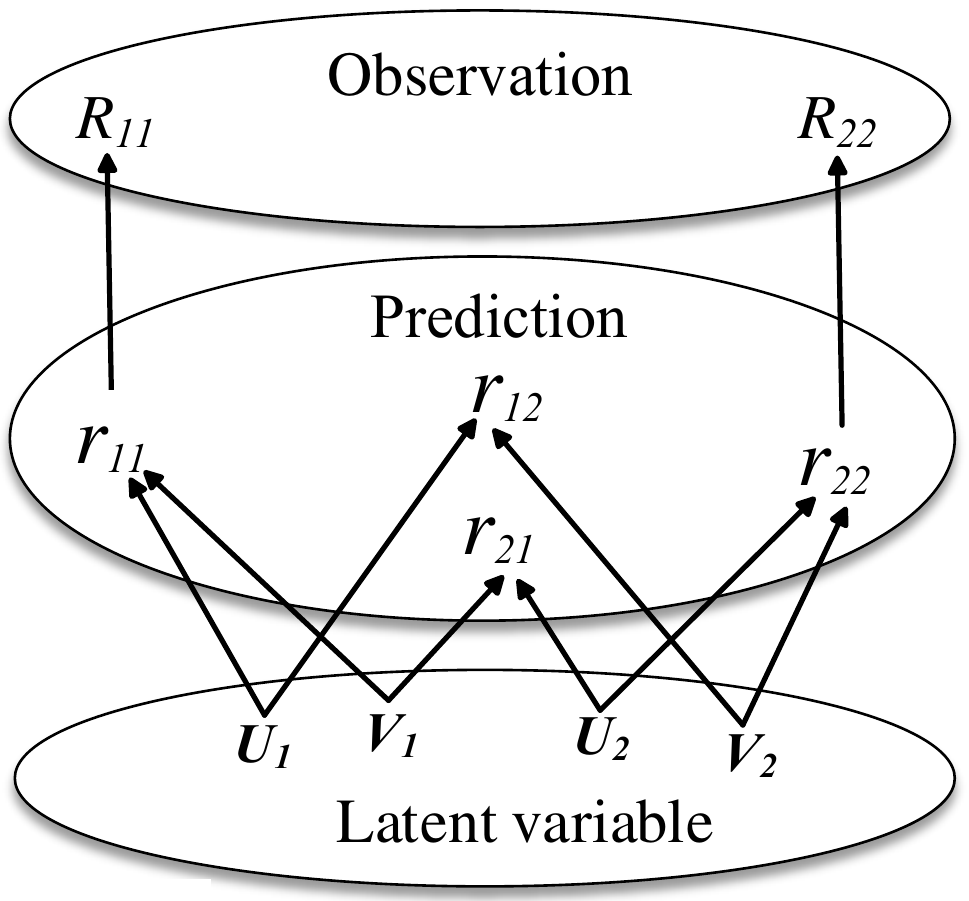}}
~~~~~
\subfigure[CGM example] { \includegraphics[width=3.5cm]{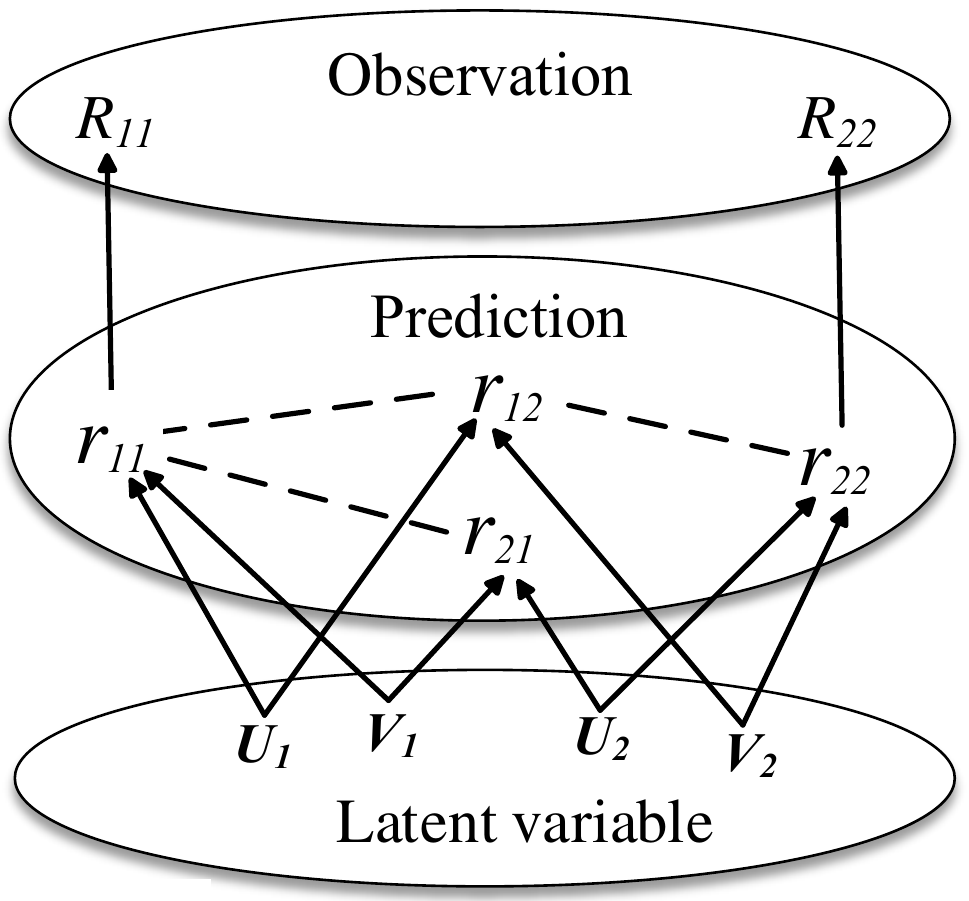}}
\caption{Toy example.}
\label{cgm-ctr}
\end{figure*}

\begin{table}
\centering
\caption{Hyperparameters notations in our model}
\label{notations}
\begin{tabular}{|c|c|}
  \hline
  \textbf{Notation} & \textbf{Meaning} \\
  \hline
  \hline
  $C^L$ & label confidence matrix \\
  \hline
  $\mu^U$ & user prior mean \\
  \hline
  $\mu^V$ & item prior mean \\
  \hline
  $\lambda_U$ & user prior confidence \\
  \hline
  $\lambda_V$ & item prior confidence \\
  \hline
  $\lambda_F$ & global smoothness degree on user graph \\
  \hline
  $\lambda_G$ & global smoothness degree on item graph \\
  \hline
  $\alpha$ & smoothness confidence decay parameter \\
  \hline
\end{tabular}
\end{table}

In this section, we propose a novel probabilistic CGM to marry SSL with LFM, which is Insight $\mathcal{I}$ (i.e., principled unification).

\subsection{Problem Statement}\label{state}

We first define our recommendation problem. 
Given a user set $\mathds{U}=\{u_1,...,u_I\}$ and an item set $\mathds{V}=\{v_1,...,v_J\}$, 
there are totally $I \times J$ ratings, among which only $\mathscr{L}$ ratings are known, and our recommendation task is to predict the unknown ratings, with the size of $\mathscr{U}=I \times J-\mathscr{L}$.

As described in Section \hyperref[intro]{\ref{intro}}, SSL and LFM are popularly used to solve the above recommendation problem, however, they have their own merits and disadvantages. 
Since we want to marry SSL with LFM, we first use a toy example to illustrate how SSL and LFM make recommendation separately. 
Let $\bm{R}$ be the U-I rating matrix, with each element $R_{ij}$ denoting the rating that $u_i$ gives to $v_j$. 
Let $\bm{r}$ be the predicted U-I rating matrix, with its real-valued rating $r_{ij}$ denoting the predicted rating of $u_i$ on $v_j$. 
Figure \ref{cgm-ctr} (a) shows a U-I rating matrix example, where we have two users ($u_1$ and $u_2$), two items ($v_1$ and $v_2$), and two known U-I rating pairs ($R_{11}$ and $R_{22}$). Our mission is to predict the two unknown U-I pairs ($r_{12}$ and $r_{21}$). 

%As described in Section \hyperref[intro]{\ref{intro}}, graph-based SSL and LFM are two popular approaches to alleviate data sparsity problem in RS. 

Graph-based SSL makes recommendation mainly based on the smoothness idea on affinity graphs. 
Since each data object in RS is a U-I rating pair, to make recommendation, SSL first needs to build a U-I pairwise affinity graph, and then propagates the known U-I rating pairs to the unknown ones. 
Let $\mathcal{G}=(\mathcal{V}, \mathcal{E})$ be a U-I \emph{pairwise affinity graph}  with nodes $\mathcal{V}$ denoting data objects, edges $\mathcal{E}$ denoting the affinitive relationship between nodes, and weight matrix between edges $P$ denoting the affinitive relation strength between nodes. 
Figure \ref{cgm-ctr} (b) shows how to adopt SSL to do recommendation, where the affinity graph $\mathcal{G}$ is shown in dash lines, and we do not show the edge weight for conciseness. Predictions will be made by propagating $R_{11}$ and $R_{22}$ to $r_{12}$ and $r_{21}$. 
That is, the purpose of SSL is rating smoothness on affinity graphs, and SSL captures the mutual dependency between ratings. 

LFM makes recommendation mainly based on the dimensionality reduction idea. 
Let $\bm{U}\in{}\mathds{R}^{K\times{}I}$ and $\bm{V}\in{}\mathds{R}^{K\times{}J}$  be the user and item latent feature matrixes, with their column vectors $\bm{U_i}$ and $\bm{V_j}$ representing the \emph{K}-dimensional latent vectors of $u_i$ and $v_j$ respectively. 
Figure \ref{cgm-ctr} (c) shows how to use LFM to do recommendation. 
LFM predicts the unknown ratings by learning user and item latent factors through regressing on the known ones. 
In this example, LFM predicts $r_{12}$ and $r_{21}$ by learning $U_1$, $U_2$, $V_1$, and $V_2$ through regressing on $R_{11}$ and $R_{22}$. 
That is, the purpose of LFM is rating regression, and LFM captures the causal dependency between latent factors and ratings. 

Although LFM and SSL have different purpose, they both are actually used to capture  different dependency between different variables. 
To marry LFM with SSL, we need a model so that such different dependency between different variables can be captured.

\subsection{Regression and Smoothness Based Chain Graph Model}\label{cfssl-a}

%Our goal is to marry LFM with SSL.
%As described in Section \hyperref[intro]{\ref{intro}}, the existing LFMs can only capture the rating regression idea, and ignore rating smoothness insight. Directly applying SSL to do recommendation captures the rating smoothness idea, but suffers from severe scalability problem. 

We identify probabilistic CGM, a powerful tool for representing the relationship between variables, to be ideal for marrying LFM with SSL. 
CGM is a combination of Bayesian network and Markov random field \cite{Koller2009Probabilistic,mccarter2014sparse}.
First, Bayesian network, i.e, directed probability graphical model, is commonly used to model the causal dependency between variables. %E.g., CTR adopts Bayesian network to model the rating regression procedure from user and item latent factors to U-I ratings \cite{wang2011collaborative}. 
Second, Markov random field, i.e., undirected probability graphical model, is used to model the mutual dependency between variables. %E.g., HF uses Gaussian Markov random field to model label smoothness objective on affinity graphs \cite{zhu2003semi}.
% Although Bayesian network and Markov random field seem conflict (directed vs. undirected), they both are used to model probability dependency between variables. 

To marry LFM with SSL, the proposed CGM should 
serve for two goals, as we described in the above example: (1) \emph{Rating regression}.
The predicted rating should be close to the observed rating; (2) \emph{Rating smoothness}. The predicted ratings should be smooth between close U-I pairs. Thus, we call our model \textbf{R}egression and \textbf{S}moothness-based \textbf{C}hain \textbf{G}raph \textbf{M}odel (RSCGM).

Our proposed RSCGM combines Bayesian network and Markov random field. 
Figure \ref{cgm-ctr} (d) shows an our proposed CGM example, where we use the directed graph (shows with solid arrows) to model rating generation and known rating regression procedures, and use the undirected graph (shows with dash lines) to model the rating smoothness constrain on an affinity graph.
RSCGM is a three layer probabilistic graphical model.
The first layer is the \emph{latent factor layer}, and its nodes are latent variables, i.e., $\bm{U}$ and $\bm{V}$. The second layer is the \emph{prediction layer}, and its nodes are the predicted ratings, i.e., $\bm{r}$. The third layer is the \emph{observation layer}, and its nodes are the observed data, i.e., $\bm{R}$. From the latent factor layer to the prediction layer, it is a Bayesian network which denotes the rating generation procedure; I.e., we use $r_{ij} = {\bm{U_i}}^T\bm{V_j}$ as the prediction of a U-I pair. The prediction layer is a Markov random field which denotes the prediction smoothness constrain. From the prediction layer to the observation layer, it is another Bayesian network which denotes the rating regression objective, i.e., the predicted rating should be close to the observed rating.

%Our proposed CGM is a combination of Bayesian network and Markov random field, and thus is able to capture rating regression and rating smoothness at the same time. The existing LFM only uses Bayesian network to capture rating regression idea and ignores rating smoothness, and that is why it suffers data sparsity to some extent.

\stitle{Joint distribution over all the variables.}We first give the joint distribution over all the variables in all the three layers. The Markov property of chain graph model \cite{lauritzen2002chain}, i.e., conditional independence relations between variables, indicates that the probability of a node on a CGM only depends on its directed neighbors. Thus, we factorize the joint distribution over all the variables  as:
\begin{equation}\label{joint-distribution}\small
\begin{split}
&P(\bm{U},\bm{V},\bm{r},\bm{R}|\bm{C^L},\bm{\mu^U},\bm{\mu^V},\lambda_U,\lambda_V)\\
% = & P(R|r,C^L,\sigma _R^2)P(r|U,V)P(U,V|\sigma _U^2,\sigma _V^2,\theta_j)\\
 \propto & P(\bm{R}|\bm{r},\bm{C^L}) P(\bm{r}|\bm{U},\bm{V}) P\left(\bm{U}|\bm{\mu^U},\lambda_U\right)P\left(\bm{V}|\bm{\mu^V},\lambda_V\right).
\end{split}
\end{equation}

%Based on the property of Dirac delta function \cite{dirac1981principles}, integrating out $r$ is equivalent to replacing $r$ with $U^TV$.

We then derive the conditional probabilities of variables in each layer.

\stitle{Latent factor layer.}Our model starts with latent factors which is also the start of the Bayesian network.
We place Gaussian priors on user latent factor:
\begin{equation}\small\label{userprior}
P\left(\bm{U}|\bm{\mu^U},\lambda_U\right)=\prod\limits_{i = 1}^I\mathcal{N}(\bm{U_i}|\bm{\mu_i^U},\lambda_U^{-1}\bm{I_K}),
\end{equation}
where $\bm{I_K}$ is a $K$-dimensional identity matrix, and $\mathcal{N}(\mu, \lambda)$ is the probability density function of the Gaussian distribution with mean $\mu$ and variance $\lambda$. Note that $\bm{\mu^U}$ is the user prior mean matrix with each column $\bm{\mu_i^U}$ denoting the mean of each user latent factor. $\bm{\mu^U}$ can be obtained from additional information. For example, \cite{jamali2010matrix} takes the average preference of a user's friends as the mean of his prior.
Similarly, we place another Gaussian prior on item factor:
\begin{equation}\small\label{itemprior}
P\left(\bm{V}|\bm{\mu^V},\lambda_V\right)=\prod\limits_{j = 1}^J\mathcal{N}(\bm{V_j}|\bm{\mu_j^V},\lambda_V^{-1}\bm{I_K}),
\end{equation}
where $\bm{\mu^V}$ is the item prior mean matrix with each column $\bm{\mu_j^V}$ denoting the mean of each item latent factor, which can also be obtained from additional information. For example, CTR \cite{wang2011collaborative} takes the topic allocation learned from the content information of an item as the mean of its prior.

\stitle{Prediction layer.}The prediction $\bm{r}$ comes from the latent factor layer, and has smoothness constrain between themselves. As a result, the probability of $\bm{r}$ depends on two independent parts: (1) variables from the first layer, i.e., $\bm{U}$ and $\bm{V}$; (2) affinitive neighbors on user and item affinity graphs.
The first part corresponds to the directed graph that comes from the latent factor layer to the prediction layer in Figure \ref{cgm-ctr}. The second part corresponds to the undirected graph in the second layer of Figure \ref{cgm-ctr}.
Based on the Markov property of chain graph model \cite{lauritzen2002chain}, we have
\begin{equation}\small\label{secondlayer}
\begin{split}
P(\bm{r}|\bm{U},\bm{V}) = \frac{1}{Z}  \phi_1(\bm{U},\bm{V},\bm{r})\phi_2(\bm{r}) \propto \phi_1(\bm{U},\bm{V},\bm{r})\phi_2(\bm{r}),
\end{split}
\end{equation}
where $Z$ is a normalizer that makes sure the probability equals to 1.

In Eq.~(\ref{secondlayer}), we define the first term $\phi_1(\bm{U},\bm{V},\bm{r})=\delta(\bm{r}-\bm{U}^T\bm{V})$ with $\delta()$ denoting the Dirac delta function \cite{dirac1981principles}, and the property of Dirac delta function indicates that integrating out $\bm{r}$ is equivalent to replacing $\bm{r}$ with $\bm{U}^T\bm{V}$, which is the rating generation procedure.
The second term $\phi_2(\bm{r})=exp\{-E\}$ is probability that constrains rating smoothness with $E$ denoting the rating smoothness energy function on affinity graphs, i.e., the confidence-aware joint-smoothness objective function that we will present in Section \hyperref[sec-energyfunction]{\ref{sec-energyfunction}}.

%Particularly, $P(r) = \prod\limits_c {{\psi _c}({r_c})} /Z$, where $Z$ is a a normalization constant given by $Z = \sum\limits_r {\prod\limits_c {{\psi _c}({r_c})} }$, and ${\psi _c}({r_c})$ is the \emph{potential functions} over the set of all possible configurations of the generated $r$ on the affinity graph. The literature commonly uses \emph{Boltzmann distribution} to express potential functions , i.e., ${\psi _c}({r_c})=exp\{-E(r_c)\}$, where $E(r_c)$ is called an \emph{energy function}, which constrains the mutually dependent relationships between nodes in each possible configuration on the undirected graph \cite{bishop2006pattern}. We will present the form the energy function in details later in Section V.

\stitle{Observation layer.}Each node in the observation layer is an observed rating of an prediction from the prediction layer, and it corresponds to the rating regression part from the prediction layer to the observation layer in Figure \ref{cgm-ctr}.
As a result, the probability of $\bm{R}$ is conditional on $\bm{r}$ in the prediction layer.
We adopt a Gaussian prior here,
%, that is, we take each rating $R_{ij}$ as a Gaussian distribution with a mean of $r_{ij}$ and a variance that allows derivation
the same as the existing LFM \cite{mnih2007probabilistic}:
\begin{equation}\small\label{conditional}
P\left(\bm{R}|\bm{r},\bm{C^L}\right)=\prod\limits_{i=1}^I\prod\limits_{j=1}^J \mathcal{N}(R_{ij}\vert{}r_{ij}, {C_{ij}^L}^{-1}) ,
\end{equation}
where $\bm{C^L}\in{}\mathds{R}^{I\times{}J}$ is a label confidence matrix with each element $C_{ij}^L$ denoting the label confidence for each U-I pair, and more details refer to \cite{wang2011collaborative}.
%Replacing $r_{ij}$ with ${U_i}^TV_j$, that is the rating generation part from the first layer to the second layer in Figure \ref{cgm-ctr} , will enable the following distribution
%\begin{equation}\small\label{regression}
%p\left(R|U,V,C,{\sigma{}}_R^2\right)=\prod\limits_{i=1}^I\prod\limits_{j=1}^J \mathcal{N}(R_{ij}\vert{}{U_i}^TV_j, %{\sigma{}}_R^2 C_{ij}^{-1}).
%\end{equation}
%Eq. (\ref{regression}) shows the rating generation and regression part from user and item latent factors to the observed ratings, corresponding to the directed graph in Figure \ref{cgm-ctr}, i.e., from the first layer to the second layer and then to the third layer.

\stitle{Posterior distribution over latent factors.}Finally, we use maximum a posteriori (MAP) probability to learn the best $\bm{U}$ and $\bm{V}$. Based on Eq.~(\ref{joint-distribution}), Eq.~(\ref{userprior}), Eq.~(\ref{itemprior}), Eq.~(\ref{secondlayer}), and Eq.~(\ref{conditional}), we have the following posterior distribution over user and item latent factors by using Bayes' theorem,
\begin{equation}\small\label{posterior}
\begin{split}
&P(\bm{U},\bm{V}|\bm{r},\bm{R},\bm{C^L},\bm{\mu^U},\bm{\mu^V},\lambda_U,\lambda_V) \\
\propto & P\left(\bm{U}|\bm{\mu^U},\lambda_U\right)P\left(\bm{V}|\bm{\mu^V},\lambda_V\right)\\
& \delta(\bm{r}-\bm{U}^T\bm{V})\phi_2(\bm{r})P(\bm{R}|\bm{r},\bm{C^L}).
\end{split}
\end{equation}

We will present how to learn the MAP distribution over the user and item latent factors in Section \ref{parameter-learning}.

\section{Realizing Smoothness}\label{sec-energyfunction}

%To find an efficient and robust way to realize smoothness in RS, 
In this section, we present the confidence-aware joint-smoothness energy function, which will enable us to obtain $\phi_2(\bm{r})$ shown in Eq.~(\ref{secondlayer}). 
%We first present the derivation from pairwise to joint smoothness, which is Insight $\mathcal{II}$ (i.e., efficient smoothness). 
%We then describe how to build user and item joint affinity graphs. 
%Finally, we propose a confidence-aware approach to realize joint smoothness, which is Insight $\mathcal{III}$ (i.e., selective smoothness).

\subsection{From Pairwise to Joint Smoothness}\label{pairwise-to-joint}

Pairwise smoothness has severe scalability problem. 
%In RS scenario, each data object is a U-I pair and its label is a U-I rating, and thus, to use the exact idea of SSL to do rating prediction in RS, we need to build a U-I pairwise affinity graph $\mathcal{G}$. Following the HF realization of SSL, the pairwise smoothness objective is shown in Eq.(\ref{ssl-pair}).
%However, it is time consuming to build such a U-I pairwise graph. 
As Section \hyperref[intro]{\ref{intro}} mentioned, building such an affinity graph requires $O{(I \times J)}^2$ for $I$ users and $J$ items, i.e., Challenge $\mathcal{II}$. 
To solve it, we propose an efficient way to perform smoothness.% which should possesss lower time complexity and similar performance with pairwise smoothness. 
%As described in Section \hyperref[intro]{\ref{intro}}, the smoothness insight is that close nodes on an affinity graph should have similar labels.

%On one hand, assuming two data object are close, e.g., $(i,j)$ and $(k,o)$, indicates that $(i,j)$ and $(k,j)$ are even closer. Because comparing with $(k,o)$, $(k,j)$ shares the same item with $(i,j)$. Similarly, $(k,j)$ and $(k,o)$ also should be even closer. $(i,j)$ and $(k,j)$ are close means that the ratings from $u_i$ and $u_k$ are close, which is the user rating smoothness insight. Similarly, $(k,j)$ and $(k,o)$ is the item rating smoothness insight. That is, pairwise smoothness indicates both user and item rating smoothness, which we call it ``joint smoothness''. 

\begin{figure*}
\centering
\subfigure[pairwise graph] { \includegraphics[width=2.2cm,height=2.1cm]{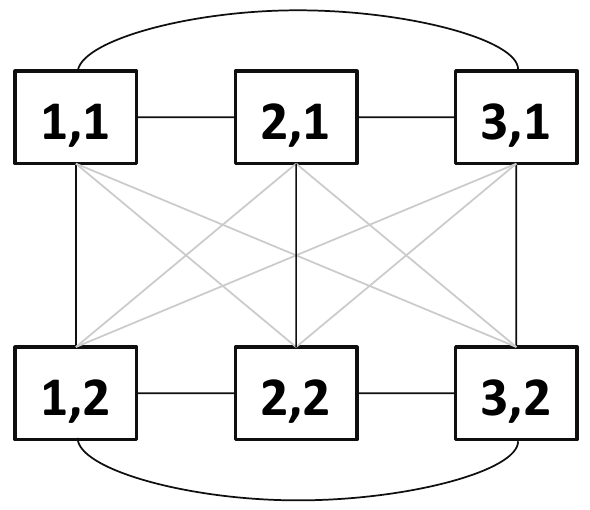}}\label{pair2joint1}~
\subfigure[remove diagonal edges] { \includegraphics[width=3.2cm,height=2.1cm]{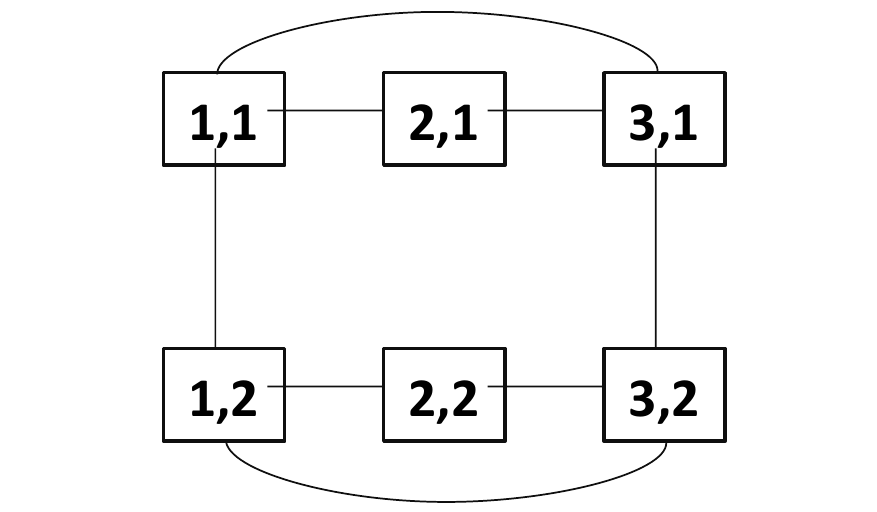}}\label{pair2joint2}~
\subfigure[merge nodes] { \includegraphics[width=2.5cm,height=2.1cm]{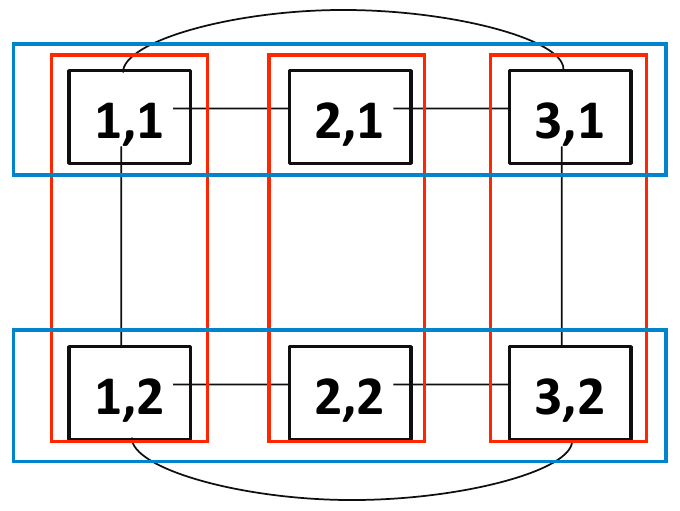}}\label{pair2joint3}~~~~~~
\subfigure[user graph] { \includegraphics[width=2.5cm,height=2.1cm]{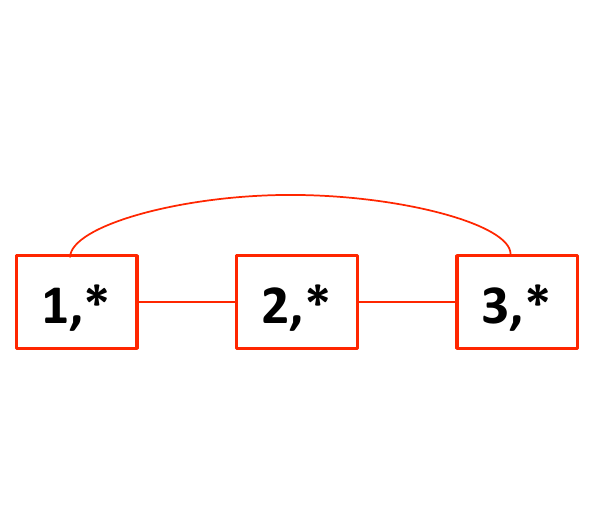}}\label{pair2joint4}
\subfigure[item graph] { \includegraphics[width=2.4cm,height=2.1cm]{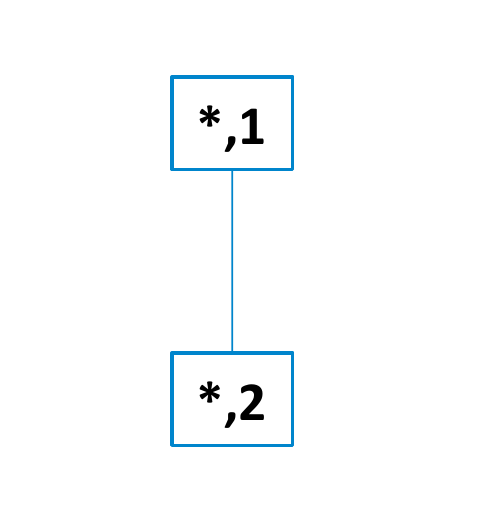}}\label{pair2joint5}
\vskip -0.15in
\caption{Pairwise to joint smoothness example.}
\label{pair2joint}
%\vskip -0.1in
\end{figure*}

In RS, each data object connects two elements, i.e., user and item. Thus, rating smoothness implies smoothness exists on two elements, i.e., user and item, and we term it ``joint smoothness''. 
We try to find the relationship between pairwise and joint smoothness. 
Figure \ref{pair2joint} shows a derivation example from pairwise to joint smoothness, where we have three users and two items, and we simply use $(i,j)$ to denote $(u_i,v_j)$ pair. 
%In this paper, we adopt HF as our energy function.
For any two pairs on $\mathcal{G}$, e.g., $(u_i,v_j)$ and $(u_k,v_o)$ ($i \ne j$ or $k \ne o$), based on HF \cite{zhu2003semi}, pairwise objective energy function in Eq.~(\ref{ssl-pair}) can be further divided into the following three terms based on different $(u_i,v_j)$ and $(u_k,v_o)$ combinations:
\begin{equation}\small\label{pair}
\begin{array}{l}
{{\mathcal L}_P} =  {{\mathcal L}_U'} + {{\mathcal L}_V'} + {{\mathcal L}_P'},
\end{array}
\end{equation}
where
\begin{equation}\small
\begin{array}{l}
{{\mathcal L}_U'} = \frac{\lambda_P}{2}\sum\limits_{i = 1}^I {\sum\limits_{k = 1}^I {\sum\limits_{j = 1}^J {{P_{ij,kj}}{{\left( {{r_{ij}} - {r_{kj}}} \right)}^2}} } }, i \ne k\\
{{\mathcal L}_V'} = \frac{\lambda_P}{2}\sum\limits_{k = 1}^I {\sum\limits_{j = 1}^J {\sum\limits_{o = 1}^J {{P_{kj,ko}}{{\left( {{r_{kj}} - {r_{ko}}} \right)}^2}} } }, j \ne o \\
{{\mathcal L}_P'} = \frac{\lambda_P}{2}\sum\limits_{i = 1}^I {\sum\limits_{k = 1}^I {\sum\limits_{j = 1}^J {\sum\limits_{o = 1}^J {{P_{ij,ko}}{{\left( {{r_{ij}} - {r_{ko}}} \right)}^2}} } } } ,i \ne k, j \ne o.
\end{array}
\end{equation}

First, we remove the diagonal edges from the pairwise graph, i.e., remove ${{\mathcal L}_P'}$ from Eq.~(\ref{pair}). 
%Because these diagonal edges do not connect nodes which have  common users or items. 
Figure \ref{pair2joint}(b) shows the result after we remove the diagonal edges from Figure \ref{pair2joint}(a). 
%Figure \ref{pair2joint} shows a pairwise to joint smoothness example, where we have three users and two items, and Figure \ref{pair2joint}(a) shows the original pairwise graph. 
%We use $\delta_U$ to denote the distance between $(u_i,v_j)$ and $(u_k,v_j)$, i.e., $\delta_U=|(u_i,v_j)-(u_k,v_j)|$, $\delta_V$ to denote the distance between $(u_k,v_j)$ and $(u_k,v_o)$, i.e., $\delta_V=|(u_k,v_j)-(u_k,v_o)|$, and $\delta_P$ to denote the distance between $(u_i,v_j)$ and $(u_k,v_o)$, i.e., $\delta_P=|(u_i,v_j)-(u_k,v_o)|$. We know that the distance between $(u_i,v_j)$ and $(u_k,v_o)$ has a upper bound of $\delta_u + \delta_v$, since $|(u_i,v_j)-(u_k,v_o)|=|(u_i,v_j)-(u_k,v_j)+(u_k,v_j)-(u_k,v_o)| \le |(u_i,v_j)-(u_k,v_j)| + |(u_k,v_j)-(u_k,v_o)|$, i.e, $\delta_P \le \delta_U + \delta_V$. In other words, the close of $(u_i,v_j)$ and $(u_k,v_j)$, and the close of $(u_k,v_j)$ and $(u_k,v_o)$ will result in the close of $(u_i,v_j)$ and $(u_k,v_o)$. Thus, we can still achieve diagonal rating smoothness after we remove diagonal edges. 
We use $\delta_U$ to denote the difference between $r_{ij}$ and $r_{kj}$, i.e., $\delta_U=|r_{ij}-r_{kj}|$, $\delta_V$ to denote the difference between $r_{kj}$ and $r_{ko}$, i.e., $\delta_V=|r_{kj}-r_{ko}|$, 
and $\delta_P$ to denote the difference between $r_{ij}$ and $r_{ko}$, i.e., $\delta_P=|r_{ij}-r_{ko}|$. We know that the difference between $r_{ij}$ and $r_{ko}$ has a upper bound of $\delta_u + \delta_v$, since 
$|r_{ij}-r_{ko}|=|r_{ij}-r_{kj}+r_{kj}-r_{ko}| \le |r_{ij}-r_{kj}| + |r_{kj}-r_{ko}|$, i.e, $\delta_P \le \delta_U + \delta_V$. 
In other words, the rating smoothness between $(u_i,v_j)$ and $(u_k,v_j)$ and the rating smoothness between $(u_k,v_j)$ and $(u_k,v_o)$ will result in the rating smoothness between $(u_i,v_j)$ and $(u_k,v_o)$.
Thus, we can still achieve diagonal rating smoothness after we remove diagonal edges.

Second, we compute the edge affinity weight. 
The U-I pairwise affinity is determined by both user and item affinity, and thus, we take pairwise affinity as a function of user affinity and item affinity. Specifically, we assume $P_{ij,ko}=F_P(F_U(u_i,u_k),F_V(v_j,v_o))$, where $0 \le F_U(u_i,u_k) \le 1$ is a function of  the similarity between $u_i$ and $u_k$, $0 \le F_V(v_j,v_o) \le 1$ is a function of the similarity between $v_j$ and $v_o$, and $0 \le F_P(F_U,F_V) \le 1$ is a function of the similarity between $(u_i,v_j)$ and $(u_k,v_o)$. We also assume $F_U(u_i,u_i)=1$ for all the users and $F_V(v_j,v_j)=1$ for all the items, which means that the similarity between a node itself is 1. For any $v_j$, since $P_{ij,kj}=F_P(F_U(u_i,u_k),1)$ holds, i.e., the weight between any $(u_i,v_j)$ and $(u_k,v_j)$ pairs is the same regardless of $v_j$, we can represent all such weights with one, i.e., $W_{ik}=P_{ij,kj}$ which is the similarity between $u_i$ and $u_k$.
Similarly, we use $S_{jo}=P_{kj,ko}$ to represent the similarity between $v_j$ and $v_o$. 

Next, since $W_{ik}=P_{ij,kj}$ and $S_{jo}=P_{kj,ko}$, we can merge nodes,  as shown in Figure \ref{pair2joint}(c). Specifically, we merge nodes $(u_i,v_j)$ for the $v_j$ into one node $u_i$ which has $J$ labels, and we merge nodes $(u_i,v_j)$ for all the $u_i$ into one node $v_j$ which has $I$ labels, as shown in Figures \ref{pair2joint}(d) and \ref{pair2joint}(e). 
That is, we decompose the pairwise graph into user and item joint graphs, and decrease the time complexity to build joint affinity graphs to $O(I^2 + J^2)$ by computing the similarity between users and items separately. Thus, ${{\mathcal L}_U'}$ and ${{\mathcal L}_V'}$ change to:
\begin{equation}\small\label{user}
{{\mathcal L}_U} = \frac{\lambda_P}{2}\sum\limits_{i = 1}^I {\sum\limits_{k = 1}^I {\sum\limits_{j = 1}^J {{W_{ik}}{{\left( {{r_{ij}} - {r_{kj}}} \right)}^2}} } },\\
\end{equation}
\begin{equation}\small\label{item}
{{\mathcal L}_V} = \frac{\lambda_P}{2}{\sum\limits_{j = 1}^J {\sum\limits_{o = 1}^J\sum\limits_{k = 1}^I  {{S_{jo}}{{\left( {{r_{kj}} - {r_{ko}}} \right)}^2}} } },\\
\end{equation}

Finally, to achieve joint smoothness, we need to leverage the effect of user and item rating smoothness. To do this, we use two parameters, i.e., $\lambda_F$ and $\lambda_G$, to control the global smoothness degree on $\mathcal{G}_1$ and $\mathcal{G}_2$.
Combining Eq.~(\ref{user}) and Eq.~(\ref{item}), we define joint smoothness objective as:
\begin{equation}\small\label{joint}
{{\mathcal L}_J} =  \frac{\lambda_F}{2}{{\mathcal L}_U} + \frac{\lambda_G}{2}{{\mathcal L}_V},
\end{equation}
where a bigger $\lambda_F$ corresponds to a higher user rating smoothness degree on $\mathcal{G}_1$ and a bigger $\lambda_G$ a higher item rating smoothness degree on $\mathcal{G}_2$. 
We will perform experiments to compare pairwise and joint smoothness in Section \hyperref[smoothnesscompare]{\ref{smoothnesscompare}}.% to show that joint smoothness outperforms pairwise smoothness in terms of both runtime and prediction.

%Now, we want to find out the relation between pairwise smoothness and user-item joint smoothness.

%Minimizing ${\cal L}_U$ or ${\cal L}_U'$ will enable us to get $R_{ij} \approx R_{kj}$, and similarly, minimizing ${\cal L}_V$ or ${\cal L}_V'$ will enable us to get ${R_{kj}} \approx {R_{ko}}$. Minimizing ${\cal L}_U$ and ${\cal L}_V$ (or ${\cal L}_U'$ and ${\cal L}_V'$) at the same time will enable us to get the similar objective as minimizing ${\cal L}_P'$, which is ${R_{ij}} \approx {R_{ko}}$, and this is exactly the idea of label propagation of SSL. In other words, ${\cal L}_P'$ is negligible when minimizing ${\cal L}_U$ and ${\cal L}_V$ at the same time. That is, joint smoothness, i.e., minimizing ${\cal L}_U$ and ${\cal L}_V$ at the same time, enable us to achieve pairwise smoothness. 

\subsection{User and Item affinity Graphs}\label{sec-energyfunction-buildgraph}

%With joint-smoothness as a way to solve Challenge $\mathcal{II}$, we 
We now explain how to build user and item affinity graphs.

\stitle{User affinity graph.}A user affinity graph should capture affinitive relationships between users, so that the ratings of close users will be similar, as we define below:

\sstitle{Definition 1:}~A \emph{user affinity graph} is an undirected weighted graph $\mathcal{G}_1=(\mathcal{V}_1, \mathcal{E}_1)$, where
(1) $\mathcal{V}_1=\mathds{U}$ is the node set; that is, each node is a user with $J$ labels, corresponding to the ratings each user gives to all them items;
(2) $\mathcal{E}_1$ is the edge set, with $W_{ik}$ denoting the weight of edge $\mathcal{E}_{ik}$; that is, $W_{ik}$ is the relationship strength between nodes $u_i$ and $u_k$, which is symmetric, i.e., $W_{ik}=W_{ki}$.

We suggest to build user affinity graph using whatever information that is available to capture the affinitive relationships between users, e.g., ratings and user social relationships. 
For example, user social relationships also indicate users' common interests, and thus can be taken as the user affinity graph. %One can use whatever information that is available to build user affinity graph.
%In this paper, we focus on using user social relationship to build user affinity graph, where there is an edge $\mathcal{E}_{ik}$ if two nodes $u_i$ and $u_k$ have social relationship, and the weight of this edge $W_{ik}=1$.

\stitle{Item affinity graph.}Similarly, an item affinity graph should capture affinitive relationships between items, so that the ratings of close items will be similar, as defined as below:

\sstitle{Definition 2:}~An \emph{item affinity graph} is an undirected weighted graph $\mathcal{G}_2=(\mathcal{V}_2, \mathcal{E}_2)$, where
(1) $\mathcal{V}_2=\mathds{V}$ is the node set; that is, each node is an item with $I$ labels, corresponding to the ratings it receives from all the users;
(2) $\mathcal{E}_2$ is the edge set, with $S_{jo}$ denoting the weight between of edge $\mathcal{E}_{jo}$; that is, $S_{jo}$ is the relationship strength between nodes $v_j$ and $v_o$, which is symmetric, i.e., $S_{jo}=S_{oj}$.

One can also build item affinity graph using whatever information that is available to capture the affinitive relationships between items, e.g., ratings and item content information. Frequently, cosine similarity, Jaccard's coefficient (JC), and Pearson correlation coefficient (PCC) are used to measure rating similarity between items \cite{ma2011recommender}.

%Since we use incomplete information to build user and affinity graphs, i.e., $\mathcal{G}_1$ and $\mathcal{G}_2$, we may not capture the real affinitive relationships between users and items. 
The quality of the affinity graphs, e.g., the reliability and density of the graph, significantly affects the recommendation result. 
The joint affinity graphs we build are unreliable, since we use rare existing information. 
Thus, we propose a confidence-aware approach to realize smoothness on them. 
We will further study the effects the affinity graph on recommendation performance during experiments. 
%Specifically, we study the effect of graph reliability and graph density on recommendation performances in Section 7.3 and Section 7.5 respectively. 

\subsection{Confidence-aware Joint-smoothness}\label{confident-pairwise-to-joint}
We finally present a confidence-aware approach to realize smoothness, which solves Challenge $\mathcal{III}$ (i.e., unreliable affinity).
%As we stated in Section \hyperref[preliminary-ssl]{\ref{preliminary-ssl}}, several types of energy functions can realize the graph-based smoothness insight. 
%In this paper, we choose the HF shown in Eq.~(\ref{ssl-pair}) as our energy function. One may use other options as well, as we presented in Section \hyperref[preliminary-ssl]{\ref{preliminary-ssl}}.

\stitle{Confidence-aware user rating smoothness.}User rating smoothness on  $\mathcal{G}_1$ constrains that close users on $\mathcal{G}_1$ have similar ratings on items. This is consistent with the reality: 
e.g., Alice is wondering which movie to watch, and she finds her close friends gave high ratings to ``The Dark Knight'', and she is likely to watch ``The Dark Knight''.

%As described in Section \hyperref[sec-energyfunction-buildgraph]{\ref{sec-energyfunction-buildgraph}}, the affinity graph is built based on the existing rare information, e.g., ratings and user social relationships, and thus is unreliable. %Thus, it  may not express the real affinitive relationships between users, which indicates that smoothness does not exist everywhere on the affinity graph. 
The built affinity graphs are unreliable, and thus full smoothness, i.e., assuming smoothness exists everywhere, will be an overly strong assumption. We identify to solve this by using selective smoothness approach. 
%Since we can not determine which edge is reliable, we can not directly select edges to constrain rating smoothness. However, as 
As described in Section \hyperref[relatedwork]{\ref{relatedwork}}, rating smoothness can be viewed as propagating  known ratings to unknown ones. 
Since the user affinity graph is unreliable, intuitively, smoothness confidence will decrease with the rating propagation length. 
%This is also consistent with reality: the confidence of the source information will decrease when it is propagating among user social network. 
%In this paper, to realize the selective smoothness insight, we only propagate the known ratings to their close neighbors through edges. 
%To do this, the edge smoothness confidence should decrease when the unknown nodes are further from known ones. 

\begin{figure}
\centering
\includegraphics[width=5cm]{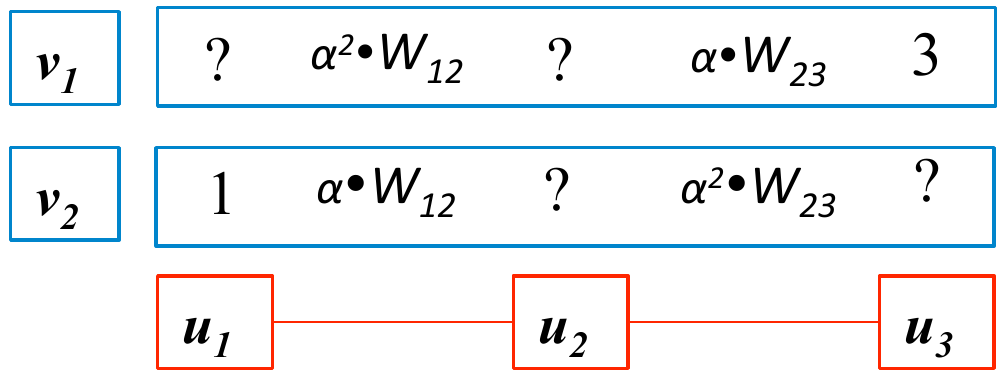}
%\vskip -0.1in
\caption{Confidence-aware user rating smoothness. %Each user has two labels, corresponding to his ratings on two items. The edge weight shows how smoothness confidence decrease with rating propagation length.
}
\label{propagation}
%\vskip -0.2in
\end{figure}

To achieve confidence-aware user rating smoothness, we propose a confidence decay parameter, i.e., $0 \le \alpha \le 1$, to control rating propagation length. 
Figure \ref{propagation} shows a user affinity graph, where we have three users and two items. 
%That is, each user is a node which has two labels, corresponding to his ratings on two items.
%, e.g., the label (3,?) of $u_3$ means that $R_{31}=3$ and $R_{32}$ is unknown. Note that $W_{ik}=P_{ij,kj}$, as we described in Section \hyperref[pairwise-to-joint]{\ref{pairwise-to-joint}}.
Although we use only one edge $W_{ik}$ to express the similarity for $(u_i,v_j)$ and $(u_k,v_j)$ pairs, there is actually a smoothness confidence between them. %, e.g., $\alpha \cdot W_{12}$ denotes the smoothness confidence between$(u_1,v_2)$ and $(u_2,v_2)$, and $\alpha^2 \cdot W_{23}$ denotes that between $(u_2,v_2)$ and $(u_3,v_2)$.
From Figure \ref{propagation}, we can see how the rating smoothness confidence decreases with propagation length. For example, the smoothness confidence between $(u_3,v_1)$ and $(u_2,v_1)$ should be bigger than that between $(u_2,v_1)$ and $(u_1,v_1)$, since the label of $(u_3,v_1)$, i.e., $R_{31}$, will first be propagated to $(u_2,v_1)$ and then to $(u_1,v_1)$.

In general, suppose we have two users ($u_i$ and $u_k$) and an item ($v_j$), we define the smoothness confidence between $(u_i,v_j)$ and $(u_k,v_j)$ pairs as:
%As we discusses in Section \hyperref[relatedwork-lfr]{II}, LFR indicates smoothness exists everywhere on the user affinity graph. However, the user affinity graph we built may not express the real affinitive relationships between users, and thus the assumption of LFR will be too strong in this scenario.
%This too strong assumption needs to propagate the known ratings to all the unknown ones, and as a result, it, on one hand, causes data noises, and on the other hand, causes unnecessary computation.
\begin{equation}\small\label{edgeconfidence}
C_{i,j,k}=\alpha^{|d_{i,j,k}|+1},
\end{equation}
where $\alpha$ is the confidence decay parameter that ranges in [0, 1], and $|d_{i,j,k}|=min\{|d_{i,j}|,|d_{k,j}|\}$ with $|d_{i,j}|$ and $|d_{k,j}|$ denoting the distances from $u_i$ and $u_k$ to a user whose rating on $v_j$ is given, respectively.
%Eq. (\ref{edgeconfidence}) indicates that the closer is an unlabelled user to a labelled user, the bigger is the edge smoothness confidence. On the contrary, with a long distance $|d|$, the smoothness confidence $\alpha^{|d|+1}$ will be close to 0, and the known label will not propagate anymore. 
%Take Figure \ref{propagation} for example, the minimum distance from $u_1$ and $u_3$ to $u_3$ is 0, and thus the original weight $W_{13}$ decay by $\alpha$. Similarly, the minimum distance from $u_1$ and $u_2$ to $u_3$ is 1, and thus the original weight $W_{12}$ decay by $\alpha^2$.

The proposed confidence-aware smoothness approach, on one hand, alleviates the overly strong assumption of fully smoothness, and, on the other hand, reduces computation. 
%On $\mathcal{G}_1$, each node $u_i$ has \emph{J} labels (both known and unknown), corresponding to the ratings he gives to each item or not.
For $u_i$ and $u_k$ on $\mathcal{G}_1$ and any $v_j$, given smoothness confidence in Eq.~(\ref{edgeconfidence}) and following Eq.~(\ref{user}), we define confidence-aware user rating smoothness energy function as:
\begin{equation}\small\label{ssl-user}
E_{U}=\mathop {\arg \min }\limits_{ \bm{U},\bm{V}} \sum\limits_{i = 1}^I \sum\limits_{k = 1}^I {\sum\limits_{j = 1}^J C_{i,j,k} {{W_{ik}}{{\left( {r_{ij} - r_{kj}} \right)}^2}}}.
\end{equation}

%Eq. (\ref{ssl-user}) shows that a user's rating on an item should be similar to his close connected neighbors. 
%The latent meaning of Eq. (\ref{ssl-user}) is that a user's rating on an item is influenced by her close friends' ratings on that item, and the stronger is their friendship, the bigger is the influence, which is consistent with reality. For instance, Alice is wondering which movie to watch, and she found most of her close friends gave high ratings to ``The Dark Knight'', and then it is very likely that Alice will think ``The Dark Knight'' is a good movie that is worth to watch.

\stitle{Confidence-aware item rating smoothness.}Similarly, item rating smoothness on $\mathcal{G}_2$ restricts that close items have similar ratings. 
%A user's rating on an item is influenced by items similar to this item, which 
This is also consistent with reality: e.g., Alice has watched movie ``The Dark Knight'' and loves it so much. She then finds ``Batman Begins'' is a similar movie, and she will probably likes ``Batman Begins'' as well.

%On $\mathcal{G}_2$, each node $v_j$ has \emph{I} labels (both known and unknown), corresponding to the ratings it received from each user or not.
%Thus, for 
Similar to user rating smoothness, for $v_j$ and $v_o$ on $\mathcal{G}_2$ and any $u_k$, following Eq.~(\ref{item}), we define confidence-aware item rating smoothness energy function as,
\begin{equation}\small\label{ssl-item}
E_{V}=\mathop {\arg \min }\limits_{\bm{U},\bm{V}} \sum\limits_{i = 1}^I \sum\limits_{j = 1}^J {\sum\limits_{o = 1}^J C_{k,j,o} {{S_{jo}}{{\left( {r_{kj} - r_{ko}} \right)}^2}}},
\end{equation}
where $C_{k,j,o}=\alpha^{|d_{k,j,o}|+1}$ has the similar meaning to Eq.~(\ref{edgeconfidence}).

%The meaning of Eq. (\ref{ssl-item}) is that a user's rating on an item is influenced by the ratings of other items similar to this item, which is also consistent with reality. For instance, Alice has watched movie ``The Dark Knight'' and loves it so much, and she found ``Batman Begins'' is a movie similar to ``The Dark Knight'', and then she will probably likes ``Batman Begins'' as well.

Following Eq. (\ref{joint}), combining Equations~(\ref{ssl-user}) \& (\ref{ssl-item}), we define  confidence-aware joint smoothness energy function as:
\begin{equation}\small\label{obj}
E_{J}=\frac{\lambda_F}{2} E_{U} + \frac{\lambda_G}{2} E_{V}.
\end{equation}

Having $E_J$ will enable us to obtain $\phi_2(\bm{r})$ shown in Eq.~(\ref{secondlayer}).

%The literature commonly uses \emph{Boltzmann distribution} to express potential functions \cite{bishop2006pattern}, i.e., ${\psi _c}({r_c})=exp\{-E(r_c)\}$, where $E(r_c)$ is the edge confidence-based joint smoothness energy function we just obtained. Thus, we have the following distribution on the undirected joint graphs is:
%\begin{equation}\scriptsize\label{joint}
%\begin{split}
%\phi(r) \propto & exp\{-\frac{1}{2\sigma _F^2} \sum\limits_{i = 1}^I \sum\limits_{k = 1}^I {\sum\limits_{j = 1}^J \alpha^{|d|+1} {{W_{ik}}{{\left( {r_{ij} - r_{kj}} \right)}^2}}} \\
%- & \frac{1}{2\sigma _G^2}\sum\limits_{i = 1}^I \sum\limits_{j = 1}^J {\sum\limits_{o = 1}^J \alpha^{|d|+1} {{S_{jo}}{{\left( {r_{ij} - r_{kj}} \right)}^2}}}\}.
%\end{split}
%\end{equation}

\section{Model Parameter Learning}\label{parameter-learning}

Based on Equations (\ref{userprior}), (\ref{itemprior}),  (\ref{secondlayer}), (\ref{conditional}), and (\ref{obj}), the posterior distribution over the user and item latent factors in Eq.~(\ref{posterior}) becomes
\begin{equation}\small
\begin{split}
& P(\bm{U},\bm{V}|\bm{r},\bm{R},\bm{C^L},\bm{\mu^U},\bm{\mu^V},\lambda_U,\lambda_V)\\
\propto & P\left(\bm{U}|\bm{\mu^U},\lambda_U\right)P\left(\bm{V}|\bm{\mu^V},\lambda_V\right)\delta(\bm{r}-\bm{U}^T\bm{V})\phi_2(\bm{r})P(\bm{R}|\bm{r},\bm{C^L})\\
= & \prod\limits_{i = 1}^I\mathcal{N}(\bm{U_i}|\bm{\mu_i^U},\lambda_U^{-1}\bm{I_K}) \times \prod\limits_{j = 1}^J\mathcal{N}(\bm{V_j}|\bm{\mu_j^V},\lambda_V^{-1}\bm{I_K})\\
\times & exp\{  - \frac{\lambda_F}{{2}}\sum\limits_{i = 1}^I {\sum\limits_{k = 1}^I {\sum\limits_{j = 1}^J {\alpha^{|d_{i,j,k}|+1}{W_{ik}}{{\left( {{r_{ij}} - {r_{kj}}} \right)}^2}} } }\\
- & \frac{\lambda_G}{2}\sum\limits_{i = 1}^I {\sum\limits_{j = 1}^J {\sum\limits_{o = 1}^J {\alpha^{|d_{k,j,o}|+1}{S_{jo}}{{\left( {{r_{ij}} - {r_{kj}}} \right)}^2}} } } \}\\
\times & \prod\limits_{i = 1}^I {\prod\limits_{j = 1}^J {{\mathcal N}({R_{ij}}|{r_{ij}}, {C_{ij}^L}^{ - 1})} }.~~~~~~~~~~~~~~~~~~~~~~~~~~~~~~~~~~~~
\end{split}
\end{equation}

Maximizing the log of the posterior probability is equivalent to minimizing the following objective function:
\begin{equation}\label{ctr-ssl}\small
\begin{split}
\mathcal{L} = & \mathop {\arg \min }\limits_{\bm{U},\bm{V}} \frac{1}{2} \sum\limits_{i = 1}^I\sum\limits_{j = 1}^J C_{ij}^L {\left(R_{ij}-{\bm{U_i}}^T\bm{V_j}\right)}^2\\
+ & \frac{{\lambda _F}}{2}\sum\limits_{i = 1}^I \sum\limits_{k = 1}^I {\sum\limits_{j = 1}^J \alpha^{|d_{i,j,k}|+1} {{W_{ik}}{{\left( {\bm{U_i}^T\bm{V_j} - \bm{U_k}^T\bm{V_j}} \right)}^2}}} \\
+ & \frac{{\lambda _G}}{2}\sum\limits_{i = 1}^I \sum\limits_{j = 1}^J {\sum\limits_{o = 1}^J \alpha^{|d_{k,j,o}|+1} {{S_{jo}}{{\left( {\bm{U_i}^T\bm{V_j} - \bm{U_i}^T\bm{V_o}} \right)}^2}}}\\
+ & \frac{{\lambda{}}_U}{2}\sum\limits_{i = 1}^I{\left(\bm{U_{i}}-\bm{\mu_i^U}\right)}^T\left(\bm{U_{i}}-\bm{\mu_i^U}\right) \\
+ & \frac{{\lambda{}}_V}{2}\sum\limits_{j = 1}^J{\left(\bm{V_{j}}-\bm{\mu_j^V}\right)}^T\left(\bm{V_{j}}-\bm{\mu_j^V}\right),\\
%-\sum\limits_{j = 1}^J\sum\limits_{n = 1}^N\log{\left(\sum_k{\theta{}}_{jk}{\beta{}}_{k,w_{jn}}\right)},~~~~~~~~~~~~~~~~~~~\\
\end{split}
\end{equation}
%where $N$ is the total number of words in the content of $v_j$, and Dirichlet prior ($\alpha{}$) is set to 1.
%In the above equation, $\lambda_U=\sigma _R^2/\sigma _U^2$, $\lambda_V=\sigma _R^2/\sigma _V^2$, $\lambda_F=\sigma _R^2/\sigma _F^2$, $\lambda_G=\sigma _R^2/\sigma _G^2$, and 
where we replace $r_{ij}$ with $\bm{U_i}^T\bm{V_j}$.
We minimize Eq.~(\ref{ctr-ssl}) by performing coordinate descent approach, that is, we fix the hyper-parameters, and iteratively optimize the user and item latent factors $\bm{U_i}$ and $\bm{V_j}$. 

Specifically, we take the gradient of $\mathcal{L}$ with respect to $\bm{U_i}$ and $\bm{V_j}$, set it to zero and get,
\begin{equation}\label{update-u}\small
\begin{split}
& \bm{U_i} \leftarrow{} \left\{ \sum\limits_{j = 1}^J \bm{V_j} C_{ij}^L \bm{V_j}^T + {\lambda{}}_U\bm{I_K} + \lambda_F \sum\limits_{k = 1}^I \sum\limits_{j = 1}^J \alpha^{|d_{i,j,k}|+1} \bm{V_j} W_{ik} \bm{V_j}^T \right.\\
 & + \left. \lambda_G \sum\limits_{j = 1}^J \sum\limits_{o = 1}^J \alpha^{|d_{k,j,o}|+1} (\bm{V_j} - \bm{V_o}) S_{jo} (\bm{V_j} - \bm{V_o})^T \right\} ^{-1}\\
 & \{\sum\limits_{j = 1}^J C_{ij}^L R_{ij} \bm{V_j}^T + {\lambda{}}_U\bm{\mu_i^U} + \lambda_F \sum\limits_{k = 1}^I \sum\limits_{j = 1}^J \alpha^{|d_{i,j,k}|+1} \bm{V_j} W_{ik} \bm{V_j}^T \bm{U_k}\},
 \end{split}
\end{equation}
\begin{equation}\label{update-v}\small
\begin{split}
& \bm{V_j} \leftarrow{} \left\{ \sum\limits_{i = 1}^I \bm{U_i} C_{ij}^L \bm{U_i}^T + {\lambda{}}_V\bm{I_K} + \lambda_G \sum\limits_{i = 1}^I \sum\limits_{o = 1}^J \alpha^{|d_{k,j,o}|+1} \bm{U_i} S_{jo} \bm{U_i}^T \right.\\
& + \left. \lambda_F \sum\limits_{i = 1}^I \sum\limits_{k = 1}^I \alpha^{|d_{i,j,k}|+1} (\bm{U_i} - \bm{U_k}) W_{ik} (\bm{U_i} - \bm{U_k})^T \right\} ^{-1}\\
& \{\sum\limits_{i = 1}^I C_{ij}^L R_{ij} \bm{U_i}^T + {\lambda{}}_V\bm{\mu_j^V} + \lambda_G \sum\limits_{i = 1}^I \sum\limits_{o = 1}^J \alpha^{|d_{k,j,o}|+1} \bm{U_i} S_{jo} \bm{U_i}^T \bm{V_o}\}.
\end{split}
\end{equation}

%Equation (\ref{update-uu}) shows how the smoothness degree parameters $\lambda_F$ and $\lambda_G$, and the confidence decay parameter $\alpha$ affect user and item latent factors.
%The bigger $\lambda_F$ and $\lambda_G$ are, the bigger influence the smoothness constrain on user and item latent factors is.
%Meanwhile, $\alpha$ controls how long we propagate the known ratings to their unlabelled neighbors. The bigger $\alpha$ is, the small the smoothness confidence decrease with the propagation length, and the longer we propagate the known ratings to their neighbors.

\begin{algorithm}[t]
\caption{Learning RSCGM}\label{learning}
\KwIn{ratings in training set $R$, user affinity graph $\mathcal{G}_1$, item affinity graph $\mathcal{G}_2$, $\alpha$, $\lambda_U$, $\lambda_V$, $\lambda_G$, $\lambda_F$}
\KwOut{user latent profile matrix $U$ and item latent profile matrix $V$}
initialize $U$ and $V$\\
\Repeat{convergence}
{\For {$r_{ij} \in R$}{
	update ${U_i}$ based on Eq. (\ref{update-u})\\
	update ${V_j}$ based on Eq. (\ref{update-v})
}
}
\Return $U$ and $V$
\end{algorithm}

We summarize the learning of RSCGM in Algorithm \ref{learning}.
After the optimal $\bm{U_i}$ and $\bm{V_j}$ are learned, they can be used to make predictions through $r_{ij} = {(\bm{U_i^*})}^T\bm{V_j^*}$.

\stitle{Computation Analysis.}We now analysis the time complexity of our model inference. %We ignore the graph build computation, since they can be done offline for only once.
From Eq.~(\ref{ctr-ssl}), we can see that the \emph{time complexity} of realizing our model is $O(\rho_R (\sum{\overline{F}} + \sum{\overline{G}}) K)$, where $\rho_R$ is the size of observed data, $\sum{\overline{F}}$ and $\sum{\overline{G}}$ are the average number of edges we need to do rating smoothness (propagation) in $\mathcal{G}_1$ and $\mathcal{G}_2$. 
%Specifically, $\sum {\bar F}  = {{\bar F}_1} + {{\bar F}_1}{{\bar F}_2} +  \cdots  + \prod\limits_d{{{\bar F}_d}}$, where ${{\bar F}_1}$ is the average number of direct neighbors of the labelled nodes on $\mathcal{G}_1$, and ${{\bar F}_d}$ is the average number of \emph{d}-th layer of neighbors of the labelled nodes on $\mathcal{G}_1$. Similarly, $\sum {\bar G}  = {{\bar G}_1} + {{\bar G}_2}^2 +  \cdots  + \prod\limits_d{{{\bar G}_d}}$, where ${{\bar G}_1}$ is the average number of direct neighbors of the labelled nodes on $\mathcal{G}_2$, and ${{\bar G}_d}$ is the average number of \emph{d}-th layer of neighbors of the labelled nodes on $\mathcal{G}_2$.

Due to the data sparsity problem, the best value of $\alpha$ is usually very small (about 0.5 in our experiments), which indicates that the known ratings will only be propagated to their close U-I pairs. With a big $d$, the smoothness degree will decay by $\alpha^d$ which will be so small that can be neglected.
%Thus, ${{\bar F}_1} > {{\bar F}_2} > {{\bar F}_d}$ with $d>2$. In our experiments, ${{\bar F}_1}\approx 10$ and ${{\bar G}_1}\approx 36$ on the \emph{MovieLens} dataset, ${{\bar F}_1}\approx 8$ and ${{\bar G}_1}\approx 30$ on the \emph{Delicious} dataset, and ${{\bar F}_1}\approx 13$ and ${{\bar G}_1}\approx 20$ on the \emph{Lastfm} dataset. 
Due to the data sparsity problem, the average number of neighbors on $\mathcal{G}_1$ and $\mathcal{G}_2$ are usually very small, which indicates that $\sum{\overline{F}} \ll \mathscr{L}$ and $\sum{\overline{G}} \ll \mathscr{L}$. 
%For example, with $\alpha=0.5$, the local smoothness confidence will decay $1/8$ times when $d=3$, and smoothness can be neglected with $d>3$. 
Besides, since $K \ll \mathscr{L}$, our algorithm scales linearly with the observed data size $\mathscr{L}$.

\section{Experiments}\label{experiments}

In this section, we present the comprehensive experiments that aimto answer four key questions:
(1) How well does our approach handle the data sparsity problem when comparing with the state-of-the-art approaches?
(2) What is the performance difference between pairwise smoothness and joint smoothness?
(3) How do user/item affinity graph and model parameters $\lambda_F$, $\lambda_G$, and $\alpha$ affect our model performance?
(4) What is the time complexity of our approach?

\subsection{RSCGM Realizations}
Our proposed RSCGM is a general model to marry SSL with LFM. As described in Section \hyperref[relatedwork]{\ref{relatedwork}}, there are mainly two kinds of LFMs in existing work.
Thus, we apply RSCGM into both of them, i.e., a basic MF (BMF) \cite{mnih2007probabilistic} and a content-based MF (CMF) \cite{wang2011collaborative}. 
%We call these two realizations BMF-RSCGM and CMF-RSCGM respectively. 
BMF assumes zero mean Gaussian on both user and item latent factors, i.e., $\mu_i^U=\mu_j^V=0$, while CMF assumes zero mean Gaussian on user latent factor and an item topic allocation mean Gaussian on item latent factor, i.e., $\mu_i^U=0$ and $\mu_j^V=\theta_j$, where $\theta_j$ is the topic allocation learned from item content information using topic modeling technique \cite{wang2011collaborative}.

\subsection{Experiment Setting}

%We then present the datasets, baseline methods, evaluation metrics, and parameter settings during our experiments.

\stitle{Datasets.}To study how RSCGM behaves under the BMF scenario, we use Movielens-10M dataset (\emph{MovieLens}) \cite{harper2016movielens} and the classic \emph{Netflix} dataset \cite{bennett2007netflix}. 
Both datasets are popular and famous, and the ratings of which are integers that rage from 0 to 5. 
To study how RSCGM behaves under the CMF scenario, we use hetrec2011-delicious-2k dataset (\emph{Delicious}) and dataset hetrec2011-lastfm-2k (\emph{Lastfm}) \cite{cantador2011second}. These datasets, as described in Table \ref{dd}, have been popularly used  \cite{purushotham2012collaborative}.
For \emph{MovieLens} and \emph{Netflix}, we compute the PCC similarity between users and items to build user and item affinity graphs.
For \emph{Delicious} and  \emph{Lastfm}, we consider the rating of a user on an item as `1' if this user has bookmarked (or listened) this item, `0' otherwise, and we take the tags of each item as its content information.
We use user social information to build user affinity graph, and compute the JC similarity between items to build item affinity graph.

\begin{table*}
\centering
\caption{Dataset description}
\label{dd}
\begin{tabular}{|c|c|c|c|c|c|c|}
  \hline
  Dataset & users & items & tags & user social relations & U-I ratings & rating density\\
  \hline
  \hline
  \emph{MovieLens} & 71,567 & 10,681 & -- & -- & 10,000,054 & 1.31\% \\
  \hline
  \emph{Netflix} & 480,189 & 17,770 & -- & -- & 100,480,507 & 1.18\% \\
  \hline
  \emph{Delicious} & 1,867 & 69,226 & 53,388 & 15,328 & 104,799 & 0.08\% \\
  \hline
  \emph{Lastfm} & 1,892 & 17,632 & 11,946 & 25,434 & 92,834 & 0.28\% \\
  \hline
\end{tabular}
\end{table*}

%During our experiments, we split each dataset into two parts---a training dataset (80\%), and a test dataset (20\%). We train all the models on the training dataset and evaluate their performance on the test dataset.
%During our experiments, we use fivefold cross validation method. That is, we divide each of the three dataset into five groups, and each group is used to test the rating prediction models that are learned from the other four groups.

\stitle{Baseline methods.}%LFM and LFR are the state-of-the-art recommendation approaches, and thus we compare RSCGM with both LFM and LFR.
For BMF scenario, we compare RSCGM with the following state-of-the-art methods:
\begin{itemize}[leftmargin=*] \setlength{\itemsep}{-\itemsep}
	\item \textbf{ICF} \cite{sarwar2001item} is an item-based collaborative filtering  approach.%, and we use PCC to compute the rating similarity between items.
	\item \textbf{SSL} \cite{ding2007learning,wang2010enhanced} is a directly application of SSL, and we use it to do rating propagation on our built item affinity graph.
	\item \textbf{BMF} \cite{mnih2007probabilistic} is a classic MF approach.
	\item \textbf{BMF-ULFR} \cite{ma2011recommender} adds user latent factor restriction (ULFR) on BMF.
	\item \textbf{BMF-UILFR} \cite{gu2010collaborative} adds both user and item latent factor restriction (UILFR) on a weighted nonnegative MF approach. To make a fair comparison, we replace the weighted nonnegative MF approach with BMF.
\end{itemize}

For CMF scenario, we compare RSCGM with the following state-of-the-art methods:

\begin{itemize}[leftmargin=*] \setlength{\itemsep}{-\itemsep}
    \item \textbf{CMF} \cite{wang2011collaborative} is a popular content information aided MF approach.
    \item \textbf{CMF-SMF} \cite{purushotham2012collaborative} adds user social factorization in CMF.
    \item \textbf{CMF-ULFR} \cite{chen2014context} adds ULFR on CMF.
    \item \textbf{CMF-UILFR} \cite{gu2010collaborative} adds UILFR on a weighted nonnegative MF approach. To make a fair comparison, we also replace the weighted nonnegative MF approach with CMF.
    \item \textbf{PACE} \cite{yang2017bridging} proposes a neural approach to bridge collaborative filtering and SSL, and we have distinguished our work from it in Section \ref{relatedwork-cgm}. Note that PACE can only support binary ratings and thus we only compare with it on \emph{Delicious} and  \emph{Lastfm} datasets. 
\end{itemize}

%Note that we compare with these LFMs and LFRs, not only because they are the most related models to ours, more importantly, it is because they are the state-of-the-art approaches to handle data sparsity problem in RS.

\stitle{Evaluation metrics.}Since ratings range from `1' to `5' on \emph{MovieLens}, we use Mean Absolute Error (\textit{MAE}) and Root Mean Square Error (\textit{RMSE}) to evaluate prediction performance \cite{ma2011recommender,chen2014context}:
\begin{equation}\label{mae}\small
MAE=\frac{\sum\limits_{(i,j) \in \tau}\vert{}R_{ij}-{r_{ij}}\vert{}}{|\tau|},
\end{equation}
\begin{equation}\label{rmse}\small
RMSE=\sqrt{\frac{1}{|\tau|}\sum\limits_{(i,j) \in \tau}{(R_{ij}-{r_{ij}})}^2},
\end{equation}
where $|\tau|$ is the number of predictions in the test dataset $\tau$.

Since ratings range in \{`0', `1'\} on \emph{Delicious} and \emph{Lastfm}, we use \emph{Precision} and \emph{Recall} to evaluate prediction performance \cite{wang2011collaborative,purushotham2012collaborative}. For each user, precision and recall are defined as follows:
\begin{equation}\footnotesize\nonumber
Precision@|M|=\frac{Number~of~items~the~user~likes~in~M}{|M|},
\end{equation}
\begin{equation}\footnotesize\nonumber
Recall@|M|=\frac{Number~of~items~the~user~likes~in~M}{Total~number~of~items~the~user~likes},
\end{equation}
where \emph{M} is the returned items. We compute the average of all the users' precision and recall in the test dataset as the final result.

\stitle{Parameter setting:} Before comparison, we first use fivefold cross validation to find the best values of $\lambda_U$ and $\lambda_V$, and then we set accordingly for other models.
For parameter $\lambda_q$ in CMF-SMF, $\lambda_f$ in BMF-ULFR and CMF-ULFR, $\lambda$ and $\mu$ in BMF-UILFR and CMF-UILFR, and $\lambda_F$ and $\lambda_G$ in our model, we find their best values in [$10^{-3}$, $10^{-2}$, $10^{-1}$, $10^0$, $10^1$, $10^2$, $10^3$]. We also search the best value of $\alpha$ in our model in [0, 1]. 
All the experiments are done on a PC by using OpenMP\footnote{http://openmp.org/} to perform parallel computing. 

%In the following section, we will compare difference model performance in Section 7.3, compare pairwise and joint smoothness in Section 7.4, study the effects of affinity graph and model parameters on model performance in 7.5, and analyze computation time of our model in Section 7.6. 

\subsection{Comparison Results and Analysis}

We compare our proposed model, i.e., RSCGM, with the state-of-the-art approaches on all the four dataset to prove the effectiveness of our approach. Our experiments are divided into two parts: comparison with the existing BMF-based models and comaprision with the existing CMF-based models. 

\subsubsection{Comparison with BMF-based models}

First, we report the experiments of comparing RSCGM and the existing BMF models conducted on \emph{MovieLens} and \emph{Netflix} datasets. Since our key insight of this paper is to use the idea of SSL to alleviate the data sparsity problem of LFM, we focus on evaluating each model's performance under different data sparsity scenerios. We use two strategies to generate datasets with different spasity: (1) rating sample, that is, we randomly sample some ratings and remove them from the original \emph{MovieLens} and \emph{Netflix} datasets, and (2) user sample, that is, we remove the users whose ratings are bigger than a certain threshold. Through rating sample strategy, we get several sub-datasets with diffferent sparsity, i.e., \emph{MovieLens-x\%}, which means we randomly
remove \emph{x\%} of the ratings from the original dataset. Similarly, through user sample strategy, we get several sub-datasets with diffferent sparsity, i.e., \emph{MovieLens-y}, which means we remove the users whose rating numbers are bigger than \emph{y}. 
Rating sample and user sample strategies can also be used together, and \emph{MovieLens-y-x\%} means we first use user sample strategy to remove the users whose rating numbers are bigger than \emph{y} and then randomly remove \emph{x\%} of the ratings from the rest dataset.

\begin{table*}
\centering\tiny
\caption{Performance comparison on rating sample \emph{MovieLens} datasets}
\label{movielens-comparison-1}
\begin{tabular}{|c|c|c|c|c|c|c|c|c|c|c|c|c|c|}
 \hline
  Datasets & \multicolumn{2}{c|}{MovieLens} & \multicolumn{2}{c|}{MovieLens-70\%} & \multicolumn{2}{c|}{MovieLens-80\%} & \multicolumn{2}{c|}{MovieLens-85\%} & \multicolumn{2}{c|}{MovieLens-90\%} & \multicolumn{2}{c|}{MovieLens-95\%} \\
  \hline
  Sparsity & \multicolumn{2}{c|}{1.31\%} & \multicolumn{2}{c|}{0.39\%} & \multicolumn{2}{c|}{0.26\%} & \multicolumn{2}{c|}{1.97\%} & \multicolumn{2}{c|}{0.14\%} & \multicolumn{2}{c|}{0.07\%} \\
  \hline
  \hline
  Metrics & MAE & RMSE & MAE & RMSE & MAE & RMSE & MAE & RMSE & MAE & RMSE & MAE & RMSE \\
  \hline
  SSL & 0.7848 & 1.1315 & 0.7987 & 1.1434 & 0.8100 & 1.1548 & 0.8273 & 1.1788 & 0.8606 & 1.2173 & 1.0407 & 1.4385 \\
  \hline
  ICF & 0.7318 & 0.9494 & 0.7966 & 1.0319 & 0.8211 & 1.0619 & 0.8398 & 1.0889 & 0.8613 & 1.1136 & 0.9087 & 1.1681 \\
  \hline
  BMF & 0.6356 & 0.8290 & 0.6780 & 0.8831 & 0.6934 & 0.9022 & 0.7072 & 0.9244 & 0.7446 & 0.9681 & 0.8254 & 1.0785 \\
  \hline
  BMF-ULFR & 0.6351 & 0.8277 & 0.6767 & 0.8805 & 0.6933 & 0.9014 & 0.7020 & 0.9178 & 0.7446 & 0.9679 & 0.8023 & 1.0376 \\
  \hline
  BMF-UILFR & \textbf{0.6351} & 0.8277 & 0.6752 & 0.8762 & 0.6916 & 0.8948 & 0.7000 & 0.9098 & 0.7415 & 0.9586 & 0.7958 & 1.0317 \\
  \hline
  RSCGM & 0.6352 & \textbf{0.8240} & \textbf{0.6702} & \textbf{0.8678} & \textbf{0.6818} & \textbf{0.8810} & \textbf{0.6889} & \textbf{0.8928} & \textbf{0.7268} & \textbf{0.9385} & \textbf{0.7899} & \textbf{1.0166} \\
  \hline
  \hline
  improv. vs. BMF & 0.07\% & 0.5\% & 1.15\% & 1.74\% & 1.67\% & 2.35\% & 2.58\% & 3.42\% & 2.39\% & 3.05\% & 4.31\% & 5.73\% \\
  \hline
\end{tabular}
\end{table*}

\begin{table*}
\centering\scriptsize
\caption{Performance comparison on user sample \emph{MovieLens} datasets}
\label{movielens-comparison-2}
\begin{tabular}{|c|c|c|c|c|c|c|c|c|c|}
 \hline
  Datasets & Sparsity & Metrics & ~~SSL~~ & ~~ICF~~ & ~~BMF~~ & BMF-ULFR~ & BMF-UILFR & ~~RSCGM~~ & improv. vs. BMF \\
  \hline
  \multirow{2}{*}{MovieLens-100} & \multirow{2}{*}{0.44\%} & MAE & 0.8362 & 0.8192 & 0.7409 & 0.7250 & 0.7232 & \textbf{0.7172} & 3.20\% \\
  \cline{3-10}
  ~ & ~ & RMSE & 1.2115 & 1.0680 & 0.9677 & 0.9435 & 0.9422 & \textbf{0.9331} & 3.58\% \\
  \hline
  \multirow{2}{*}{MovieLens-75} & \multirow{2}{*}{0.37\%} & MAE & 0.8706 & 0.8316 & 0.7715 & 0.7531 & 0.7481 & \textbf{0.7442} & 3.53\% \\
  \cline{3-10}
  ~ & ~ & RMSE & 1.2606 & 1.0850 & 1.0070 & 0.9844 & 0.9737 & \textbf{0.9555} & 5.11\% \\
  \hline
  \multirow{2}{*}{MovieLens-50} & \multirow{2}{*}{0.30\%} & MAE & 0.8958 & 0.8616 & 0.8105 & 0.7781 & 0.7724 & \textbf{0.7649} & 5.62\% \\
  \cline{3-10}
  ~ & ~ & RMSE & 1.2895 & 1.1244 & 1.0543 & 1.0118 & 1.0029 & \textbf{0.9858} & 6.50\% \\
  \hline
\end{tabular}
\end{table*}

\begin{table*}
\centering\scriptsize
\caption{Performance comparison on user and rating sample \emph{Netflix} datasets}
\label{netflix-comparisona}
\begin{tabular}{|c|c|c|c|c|c|c|c|c|c|}
 \hline
  Datasets & Sparsity & Metrics & ~~SSL~~ & ~~ICF~~ & ~~BMF~~ & BMF-ULFR~ & BMF-UILFR & ~~RSCGM~~ & improv. vs. BMF \\
  \hline
  \multirow{2}{*}{Netflix-200-70\%} & \multirow{2}{*}{0.14\%} & MAE & 0.8846 & 0.8639 & 0.7573 & 0.7569 & 0.7532 & \textbf{0.7477} & 1.27\% \\
  \cline{3-10}
  ~ & ~ & RMSE & 1.2968 & 1.1147& 0.9856 & 0.9827 & 0.9710 & \textbf{0.9534} & 3.26\% \\
  \hline
  \multirow{2}{*}{Netflix-200-80\%} & \multirow{2}{*}{0.09\%} & MAE & 0.9187 & 0.8879 & 0.7738 & 0.7708 & 0.7673 & \textbf{0.7625} & 1.45\% \\
  \cline{3-10}
  ~ & ~ & RMSE & 1.3418 & 1.1463 & 1.0058 & 0.9975 & 0.9814 & \textbf{0.9691} & 3.65\% \\
  \hline
  \multirow{2}{*}{Netflix-200-90\%} & \multirow{2}{*}{0.05\%} & MAE & 1.0264 & 0.9131 & 0.7943 & 0.7939 & 0.7882 & \textbf{0.7750} & 2.43\% \\
  \cline{3-10}
  ~ & ~ & RMSE & 1.4781 & 1.1758 & 1.0347 & 1.0118 & 1.0009 & \textbf{0.9842} & 4.89\% \\
  \hline
\end{tabular}
\end{table*}

During our experiments, we use five-fold cross validation method. 
Table \ref{movielens-comparison-1} shows the \emph{MAE} and \emph{RMSE} comparison result on different rating sample \emph{MovieLens} datasets.
Table \ref{movielens-comparison-2} shows the \emph{MAE} and \emph{RMSE} comparison result on different user sample \emph{MovieLens} datasets.
Table \ref{netflix-comparisona} shows the \emph{MAE} and \emph{RMSE} comparison result on different user and rating sample \emph{MovieLens} datasets.
 From them, we have the following observations:
\begin{itemize}[leftmargin=*] \setlength{\itemsep}{-\itemsep}
    \item The latent factor models significantly outperforms SSL and ICF, due to its dimensionality reduction technique.
    \item With the increase of the data sparsity degree, the performance of all the models decrease. This is the standard data sparsity problem. 
    \item our approach always achieves the best performance. Besides, no matter which kind of sample strategy we use, either rating sample, user sample, or both user and rating sample, there is an obvious trend: the sparser the dataset is, the bigger improvement of our model against the comparison methods. Take the experimental results on different sampled \emph{Netflix} datasets for example, the \emph{RMSE} improvement of our approach over BMF are 3.26\%, 3.65\%, and 4.89\% on \emph{Netflix-200-70\%}, \emph{Netflix-200-80\%}, and \emph{Netflix-200-90\%} respectively.
%However, LFRs' performance are limited due to their overly strong assumption that connected users or items tend to share similar latent factors. 
%This overly strong assumption of LFR fails particularly when affinity graphs are unreliable in the data sparsity scenario. This is why CMF-ULFR and CMF-UILFR only slightly outperforms CMF, while 

%Thus, our approach always achieves the best performance.
\end{itemize}

\subsubsection{Comparison with CMF-based models}

\begin{figure*}[h]
\centering
%\subfigure[\emph{Delicious} K=100] { \includegraphics[width=5.5cm]{figures/lastfm100}}
%\subfigure[\emph{Delicious} K=150] { \includegraphics[width=5.5cm]{figures/lastfm150}}
%\subfigure[\emph{Delicious} K=200]{ \includegraphics[width=5.5cm]{figures/lastfm200}}\\
\subfigure[\emph{Delicious} K=100] { \includegraphics[width=4.5cm,height=3.7cm]{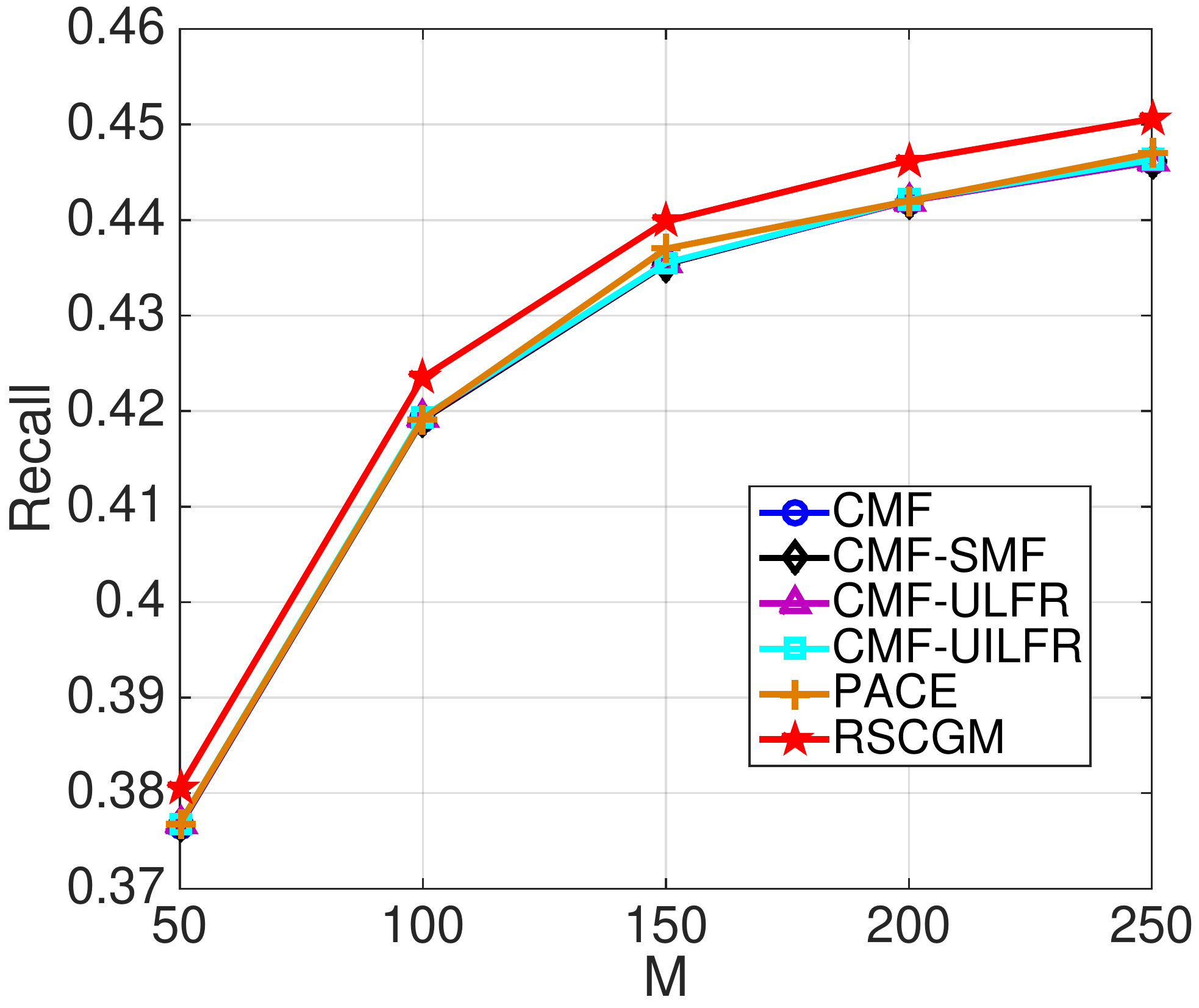}}~~~~~
\subfigure[\emph{Delicious} K=150]{ \includegraphics[width=4.5cm,height=3.7cm]{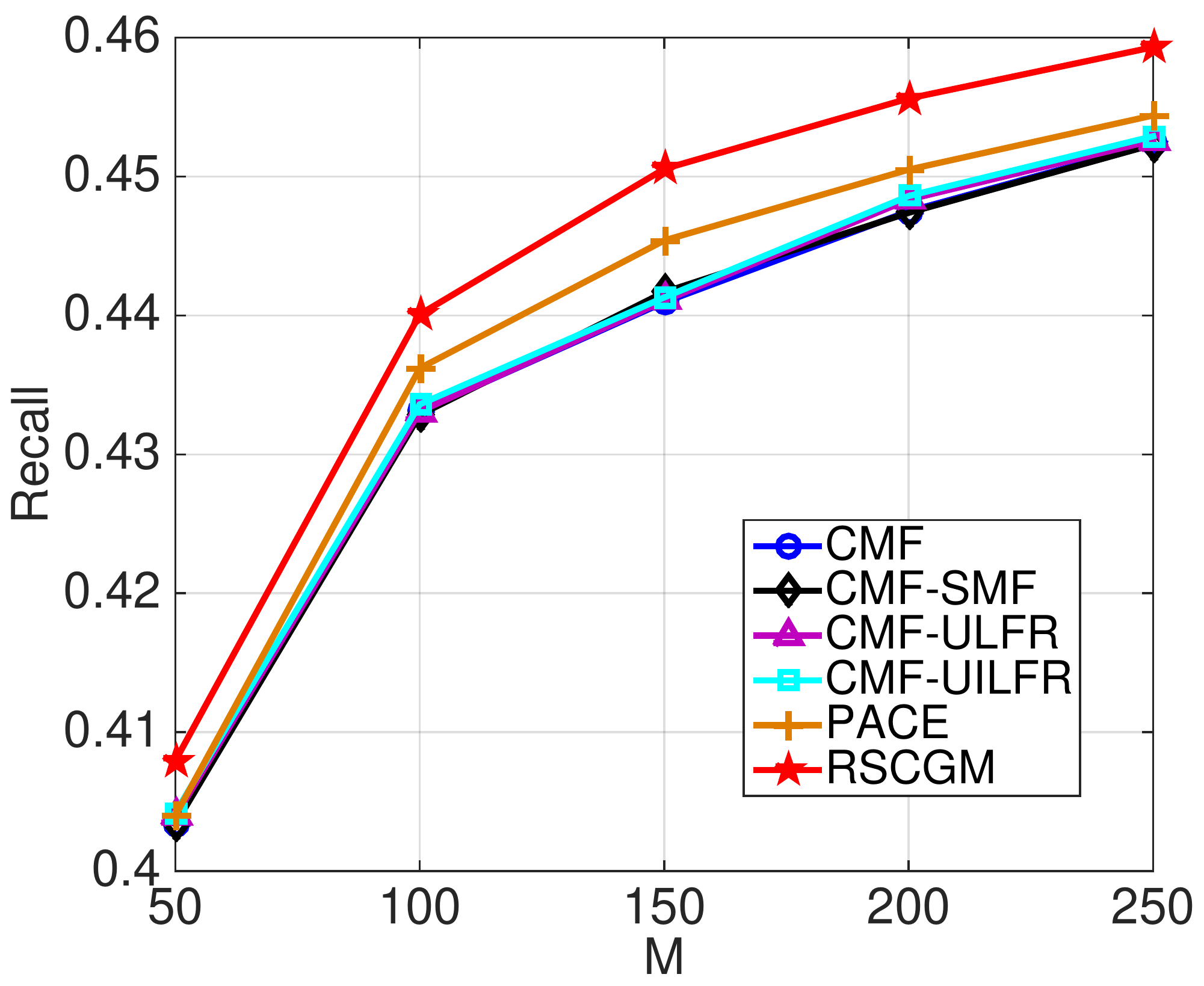}}~~~~~
\subfigure[\emph{Delicious} K=200]{ \includegraphics[width=4.5cm,height=3.7cm]{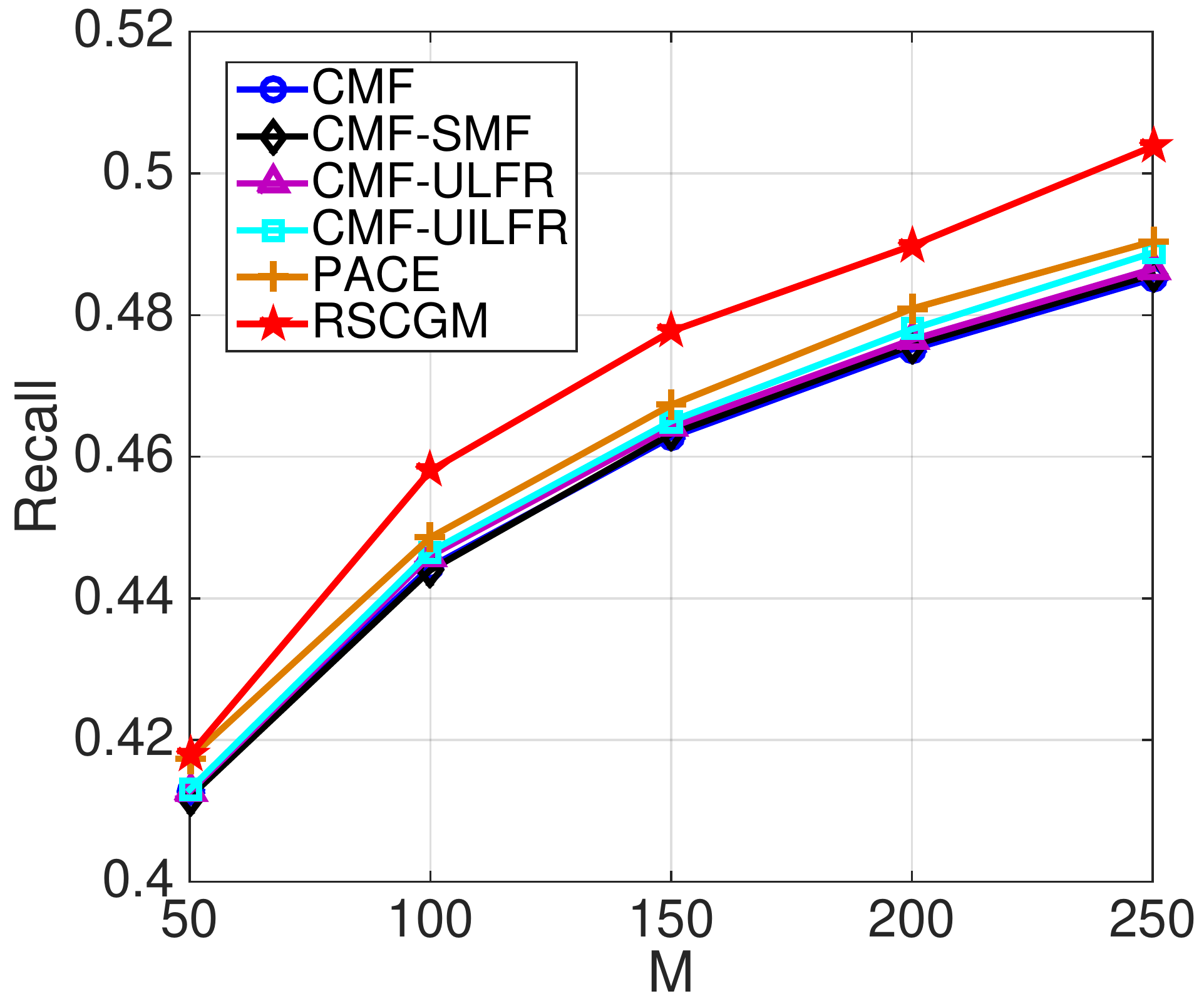}}\\
%\subfigure[\emph{Lastfm} K=100] { \includegraphics[width=5.5cm]{figures/lastfm-precision-100}}
%\subfigure[\emph{Lastfm} K=150]{ \includegraphics[width=5.5cm]{figures/lastfm-precision-150}}
%\subfigure[\emph{Lastfm} K=200]{ \includegraphics[width=5.5cm]{figures/lastfm-precision-200}}\\
\subfigure[\emph{Lastfm} K=100]{ \includegraphics[width=4.5cm,height=3.7cm]{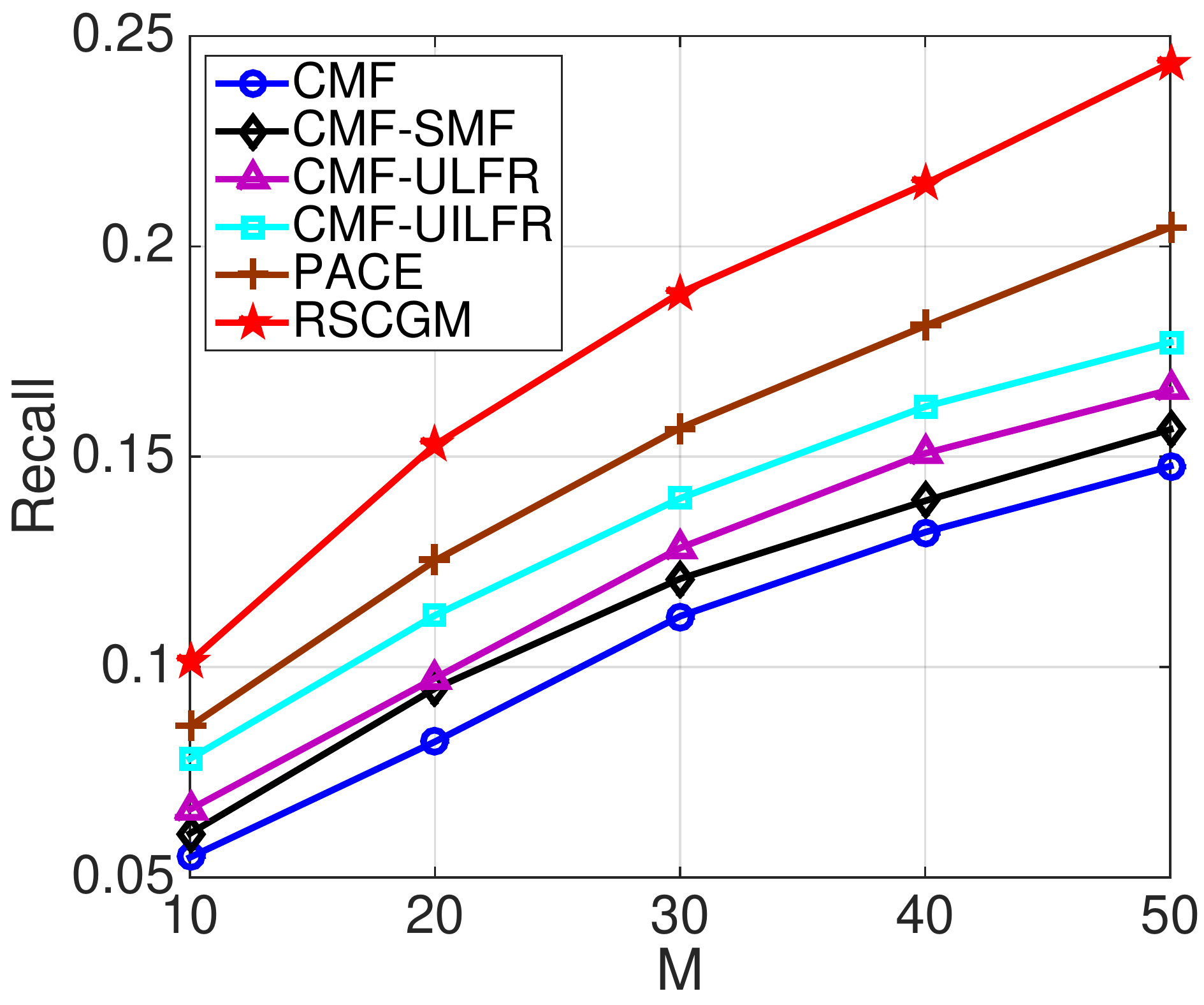}}~~~~~
\subfigure[\emph{Lastfm} K=150]{ \includegraphics[width=4.5cm,height=3.7cm]{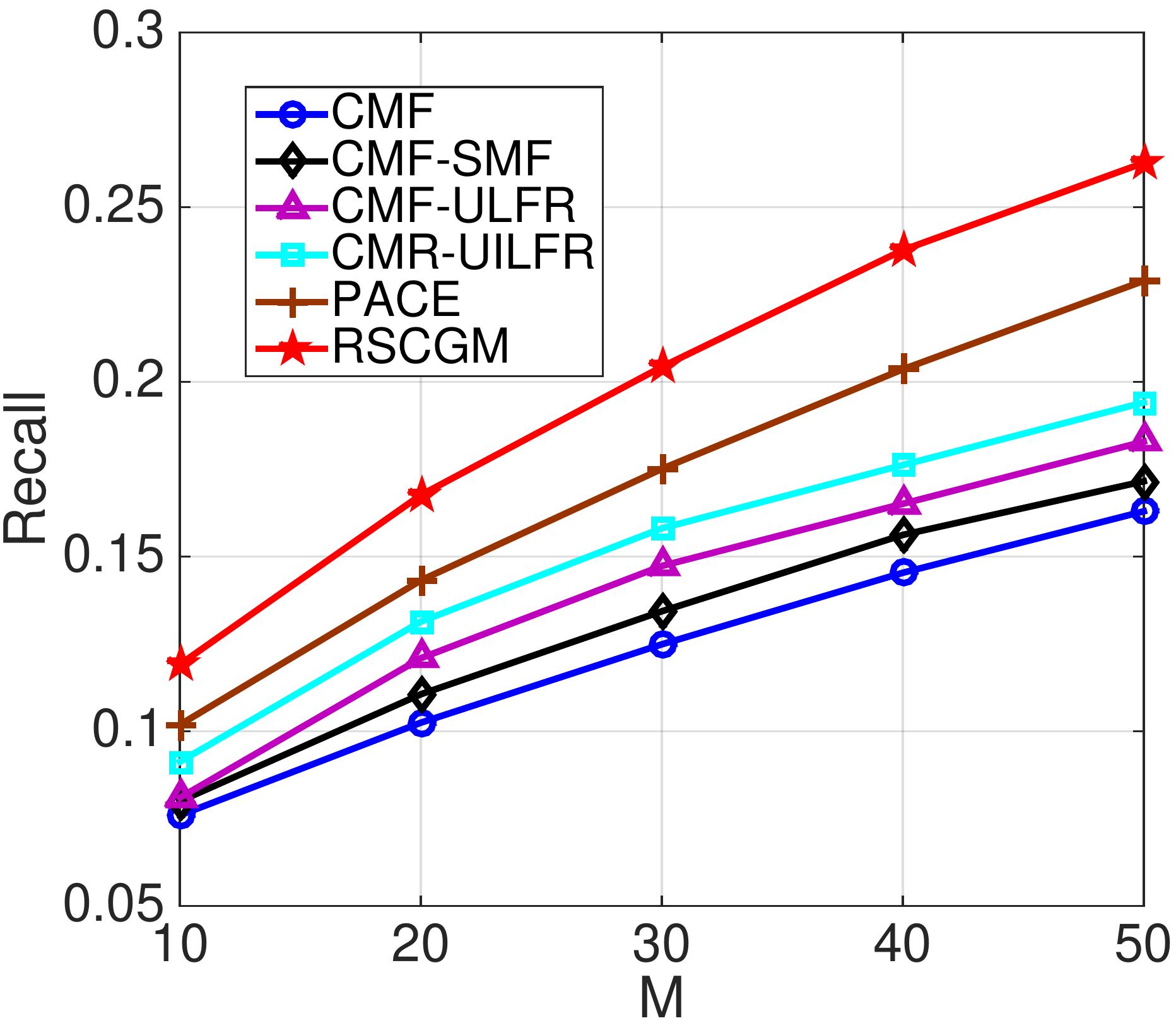}}~~~~~
\subfigure[\emph{Lastfm} K=200] { \includegraphics[width=4.5cm,height=3.7cm]{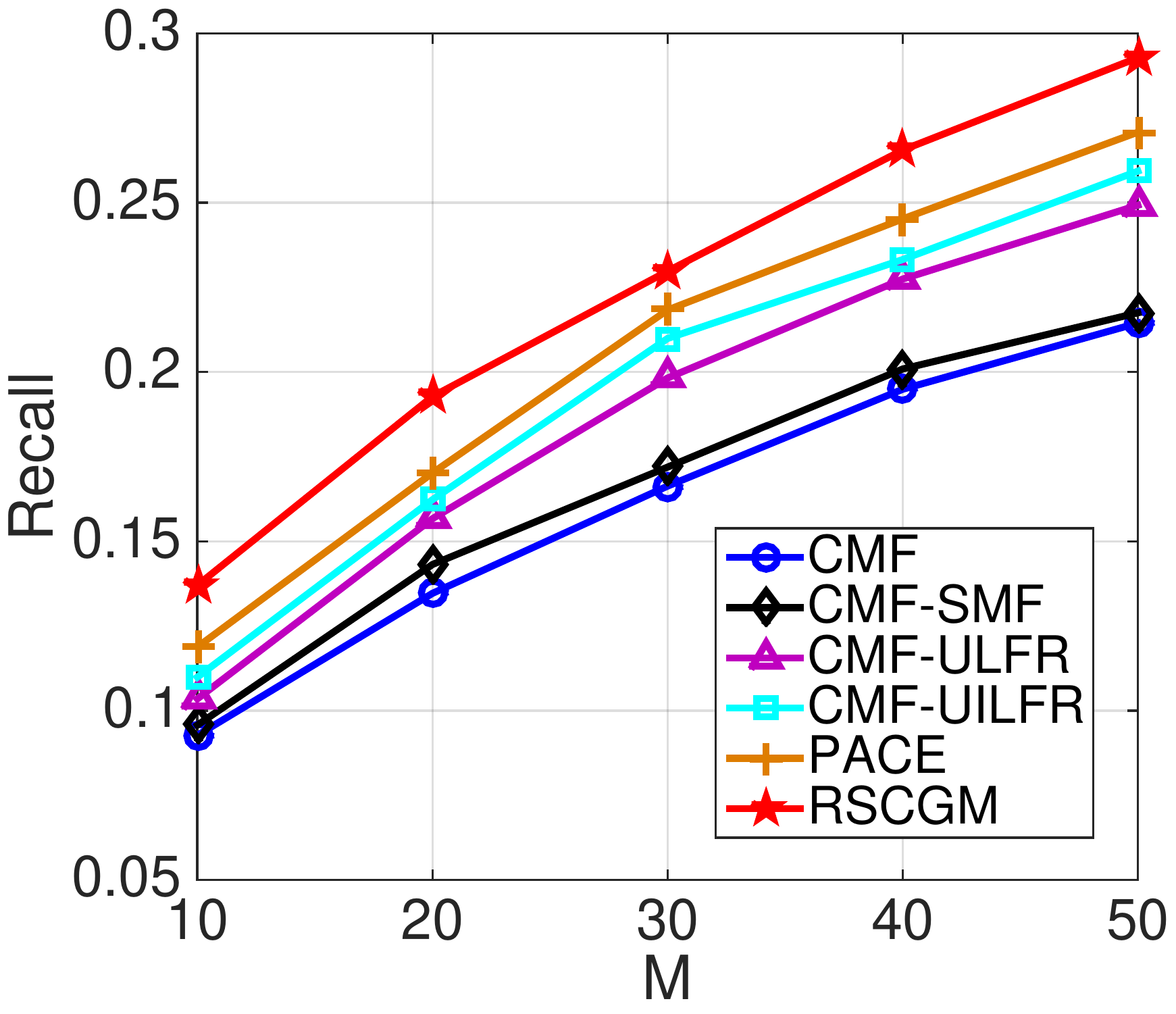}}
%\vskip -0.15in
\caption{\emph{Recall} comparison of each method on \emph{Delicious} and \emph{Lastfm}.}
\label{ctr-compare}
%\vskip -0.1in
\end{figure*}

\begin{figure*}
\centering
\subfigure[\emph{Lastfm-20\%} K=100] { \includegraphics[width=4.5cm,height=3.7cm]{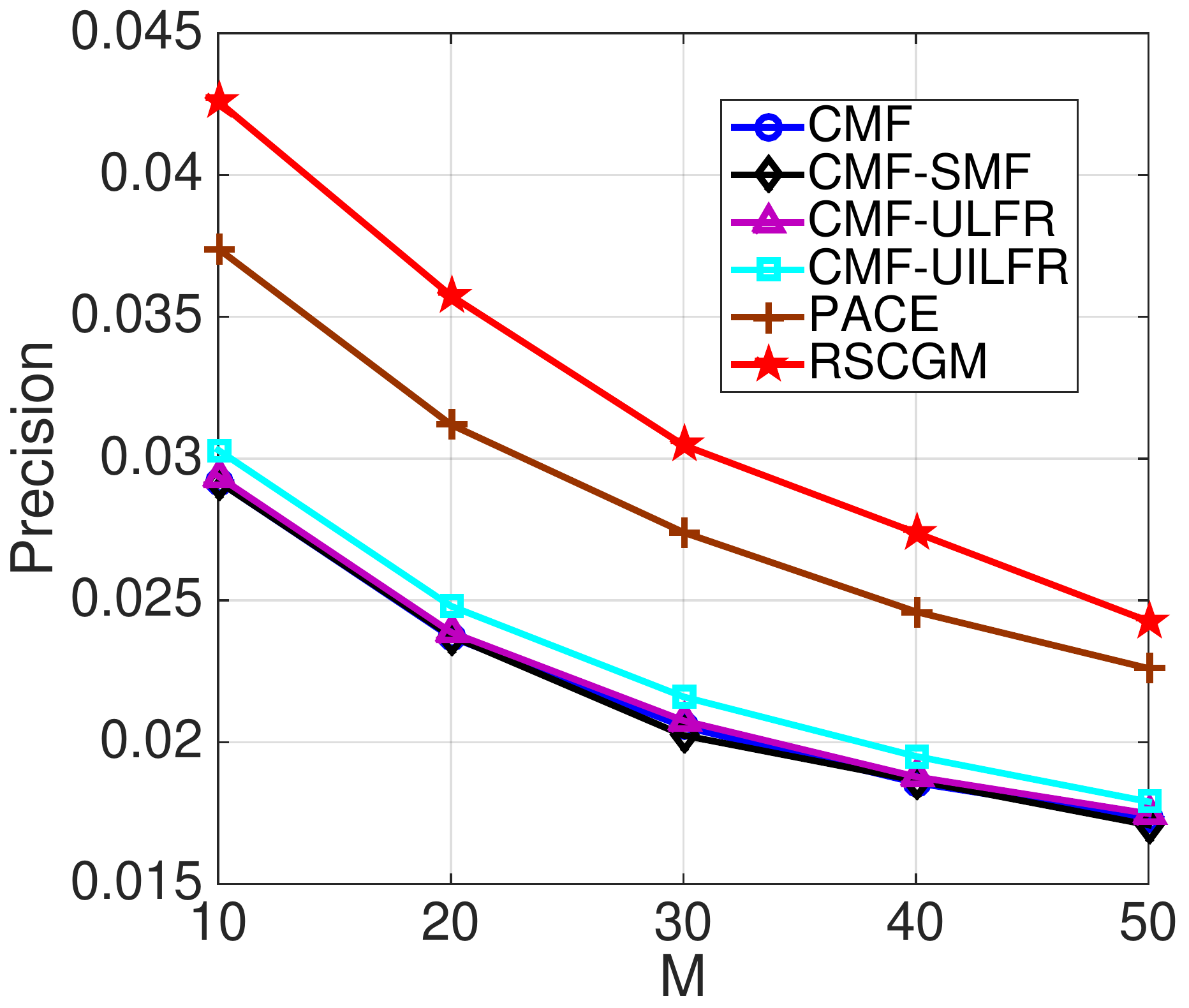}}~~~~~
\subfigure[\emph{Lastfm-20\%} K=150]{ \includegraphics[width=4.5cm,height=3.7cm]{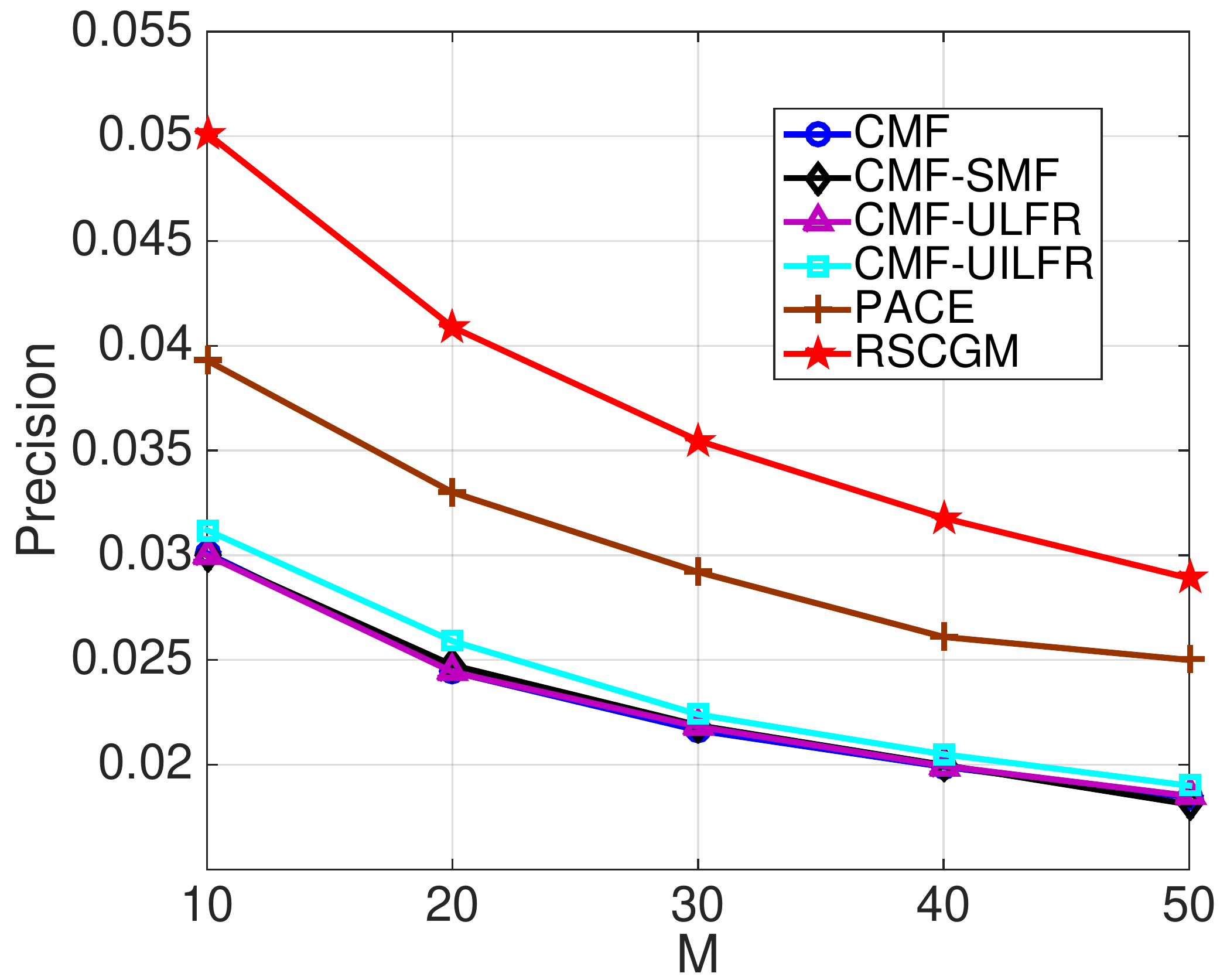}}~~~~~
\subfigure[\emph{Lastfm-20\%} K=200]{ \includegraphics[width=4.5cm,height=3.7cm]{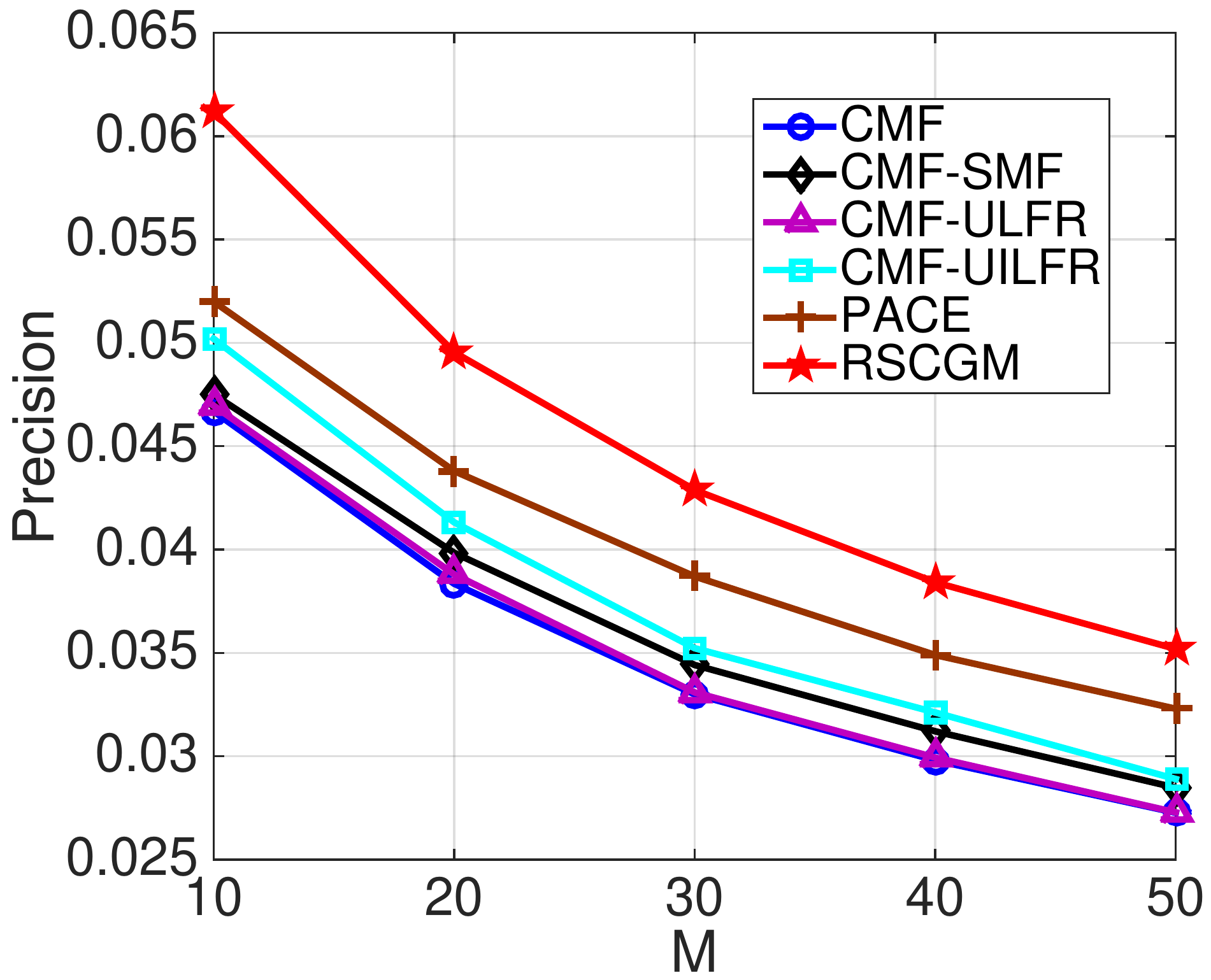}}\\
%\subfigure[\emph{Lastfm20} K=100] { \includegraphics[width=4.1cm,height=3.4cm]{figures/lastfm20-recall-100}}
\subfigure[\emph{Lastfm-50\%} K=100] { \includegraphics[width=4.5cm,height=3.7cm]{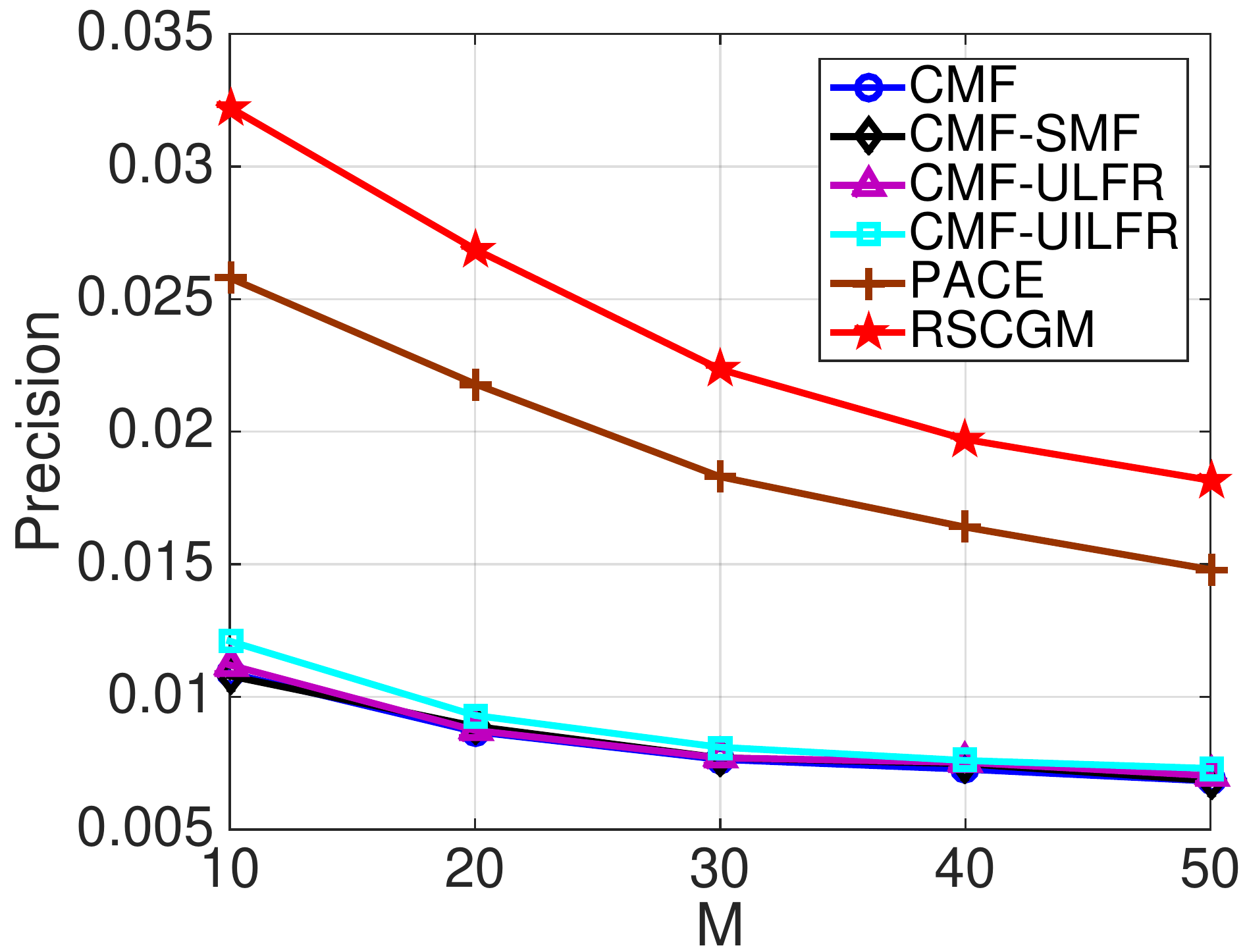}}~~~~~
%\subfigure[\emph{Lastfm50} K=100]{ \includegraphics[width=4.1cm,height=3.4cm]{figures/lastfm50-recall-100}}\\
%\subfigure[\emph{Lastfm20} K=150] { \includegraphics[width=4.1cm,height=3.6cm]{figures/lastfm20-precision-150}}
%\subfigure[\emph{Lastfm20} K=150]{ \includegraphics[width=4.1cm,height=3.6cm]{figures/lastfm20-recall-150}}
\subfigure[\emph{Lastfm-50\%} K=150]{ \includegraphics[width=4.5cm,height=3.7cm]{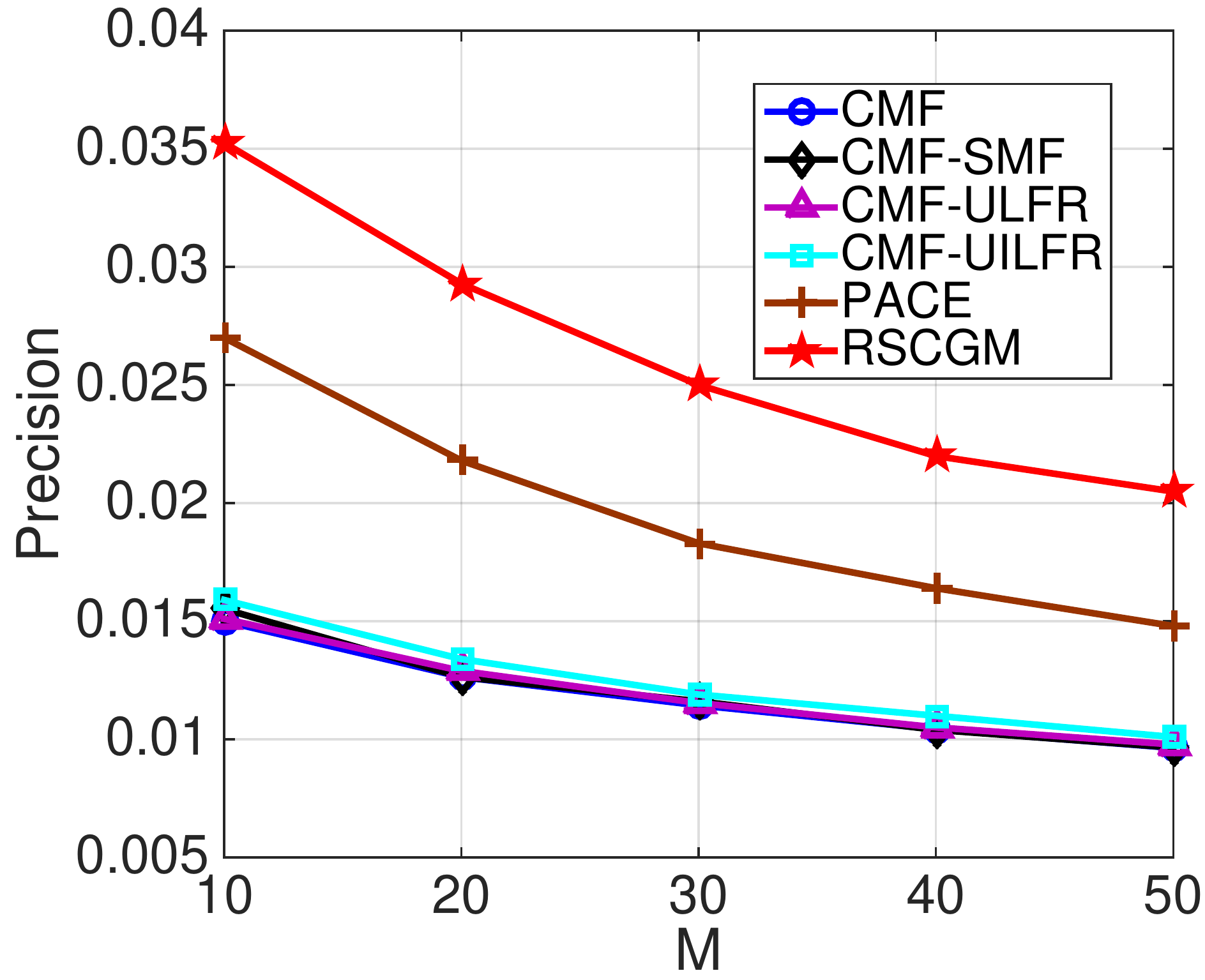}}~~~~~
%\subfigure[\emph{Lastfm50} K=150]{ \includegraphics[width=4.1cm,height=3.6cm]{figures/lastfm50-recall-150}}\\
%\subfigure[\emph{Lastfm20} K=200]{ \includegraphics[width=4.1cm,height=3.4cm]{figures/lastfm20-precision-200}}
%\subfigure[\emph{Lastfm20} K=200]{ \includegraphics[width=4.1cm,height=3.4cm]{figures/lastfm20-recall-200}}
\subfigure[\emph{Lastfm-50\%} K=200]{ \includegraphics[width=4.5cm,height=3.7cm]{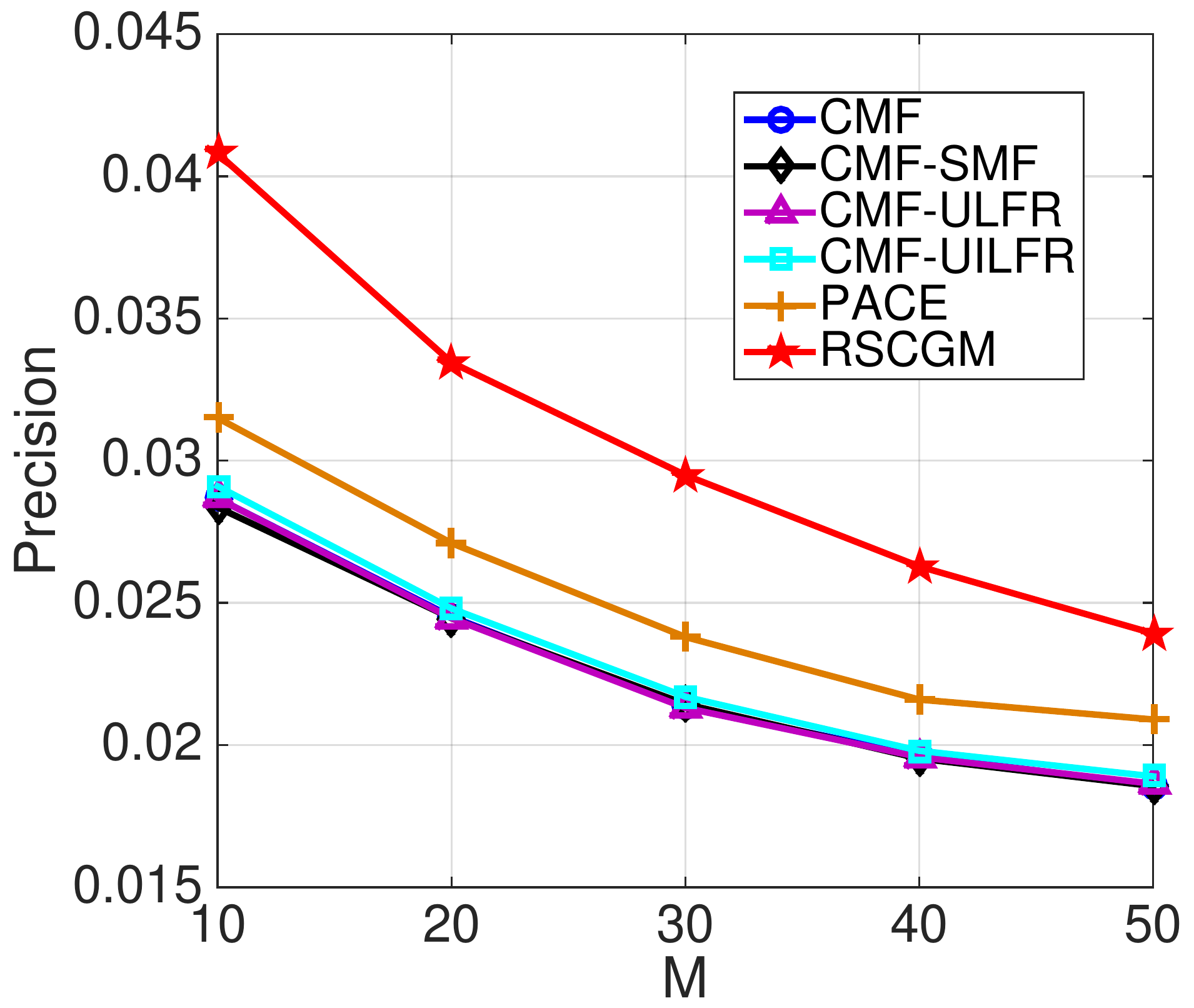}}
%\subfigure[\emph{Lastfm50} K=200] { \includegraphics[width=4.1cm,height=3.4cm]{figures/lastfm50-recall-200}}
%\vskip -0.15in
\caption{\emph{Precision} comparison of each method on \emph{Lastfm-20\%} and \emph{Lastfm-50\%}.}
\label{ctr-compare-sparsity}
%\vskip -0.2in
\end{figure*}

Next, we report the experiments of comparing RSCGM and the existing CMF models conducted on \emph{Delicious} and \emph{Lastfm} datasets. 
Note that the original \emph{Delicious} dataset is already quite sparse (0.08\%) and \emph{Lastfm} dataset is relative dense. 
We use rating sample strategy on \emph{Lastfm} to generate datasets with different sparsity degress, and we get \emph{Lastfm20} and \emph{Lastfm50} whose sparsity are 0.22\% and 0.14\%, respectively. 
Figure \ref{ctr-compare} shows the \emph{Recall} performance on \emph{Delicious} and \emph{Lastfm}, and we omit the \emph{Precision} performance since it is similar to \emph{Recall}.
Figure \ref{ctr-compare-sparsity} shows the \emph{Precision} performance for each model on different sparsity of datasets, and here we omit the \emph{Recall} performance since it is similar to \emph{Precision}. From it, we get:
\begin{itemize}[leftmargin=*] \setlength{\itemsep}{-\itemsep}
    \item With the increase of the data sparsity degree, the performance of all the models decrease. This is the standard data sparsity problem.
    \item With the increase of the latent factor dimensionality $K$, the performances of all the models increase, since a bigger $K$ will represent a better latent factor. However, as we will analysis later, it comes with price of higher model training time.
    \item CMF-SMF and CMF-ULFR slightly outperform CMF due to the additional user social information. Meanwhile, CMF-UILFR adopts additional user and item graph information and achieves better performance. Moreover, PACE behaves as a constant runner-up due to its deeper network to handle rating and affinity graph information. However, PACE uses graph embedding technique to learn user/item latent factors, which assumes graphs are strongly reliable. Thus, the performance of PACE is limited in graph unreliable situations, although it enjoys great performance in POI recommendation scenario. 
    \item With the increase of the data sparsity degree, CMF, CMF-SMF, and LFRs (CMF-ULFR and CMF-UILFR) tend to have more similar performance. This is because LFRs' performance are limited due to their overly strong assumption that connected users or items tend to share similar latent factors. This overly strong assumption of LFR fails particularly when affinity graphs are unreliable in the data sparsity scenario. 
    \item Our model (RSCGM) significantly outperforms PACE and consistently achieves the best performance among all the approaches on all the datasets. Because the marriage of SSL and LFM and the realization of the confidence-aware joint-smoothness.
    %\item The sparser the dataset is, the bigger improvement of our model against the three comparison methods. Take $K=200$ for example, the average \emph{Precision} improvement of our approach over other models are 9.82\%, 26.12\%, and 35.58\% on \emph{Lastfm}, \emph{Lastfm20}, and \emph{Lastfm50} respectively. This is because our approach benefits from confidence-aware rating smoothness we use on the affinity graphs, which alleviates the overly strong assumption of LFR.
    %the average \emph{Recall} improvement of our approach over other models are 28.27\%, 35.13\%, and 48.83\% on \emph{Lastfm}, \emph{Lastfm20}, and \emph{Lastfm50} respectively.

\end{itemize}

\subsection{Compare Pairwise and Joint Smoothness}\label{smoothnesscompare}

We then study the runtime performance (including graph build time and model inference time) and prediction performance between pairwise and joint smoothness. 
As analysied in Section \hyperref[pairwise-to-joint]{\ref{pairwise-to-joint}}, building a pairwise affinity graph is quite time-consuming, i.e., Challenge $\mathcal{II}$. 
To compare pairwise and joint smoothness, we use hetrec2011-movielens-2k dataset (\emph{MovieLens2K}) \cite{cantador2011second}, which is a relative small dataset. 
\emph{MovieLens2K} contains 2,113 users, 10,197 items, and 855,598 U-I rating pairs. 
We then use different smoothness objective function, i.e., the pairwise smoothness shown in Eq.~(\ref{ssl-pair}) and the joint smoothness shown in Eq.~(\ref{joint}), in our proposed model. We term joint smoothness ``joint'' here for simplification. We also use two kinds of approaches to build the pairwise affinity graph, i.e.,  (1) ``pairwise1'': $P_{ij,ko} = W_{ij} * S_{ko}$, and (2) ``pairwise2'': $P_{ij,ko} = min\{W_{ij},S_{ko}\}$. We finally compare their runtime and prediction differences on \emph{MovieLens2K}. The results are shown in Table \ref{smoothness-compare}, where we set $K=6$.  As we analyzed in Section \hyperref[pairwise-to-joint]{\ref{pairwise-to-joint}},  the runtime of joint smoothness is significantly shorter than pairwise smoothness. Besides, the prediction performance of joint smoothness also outperforms pairwise smoothness. This is because joint smoothness uses two parameters ($\lambda_F$ and $\lambda_G$) to control the global smoothness degree on $\mathcal{G}_1$ and $\mathcal{G}_2$ separately. In contrast, pairwise smoothness uses only one parameter ($\lambda_P$) to control the global smoothness degree on $\mathcal{G}$. Thus, joint smoothness can leverage smoothness degree more delicately on affinity graphs.

\begin{table}
\centering\small
\caption{Performance comparison between pairwise and joint smoothness on \emph{Movielens2K}.}
\label{smoothness-compare}
\begin{tabular}{|c|c|c|c|}
 \hline
  Metric & MAE & RMSE & runtime (seconds) \\
  \hline
  \hline
  pairwise1 & 0.7071 & 0.9170 & 337 \\
  \hline
  pairwise2 & 0.6889 & 0.8916 & 157 \\
  \hline
  joint & \textbf{0.6185} & \textbf{0.8188} & \textbf{5}\\
  \hline
\end{tabular}
%\vskip -0.10in
\end{table}

\subsection{Effect of Affinity Graph and Model Parameters}\label{parametereffect}

\begin{figure}
\centering
\subfigure [\emph{Delicious} $K=200$]{ \includegraphics[width=3.7cm]{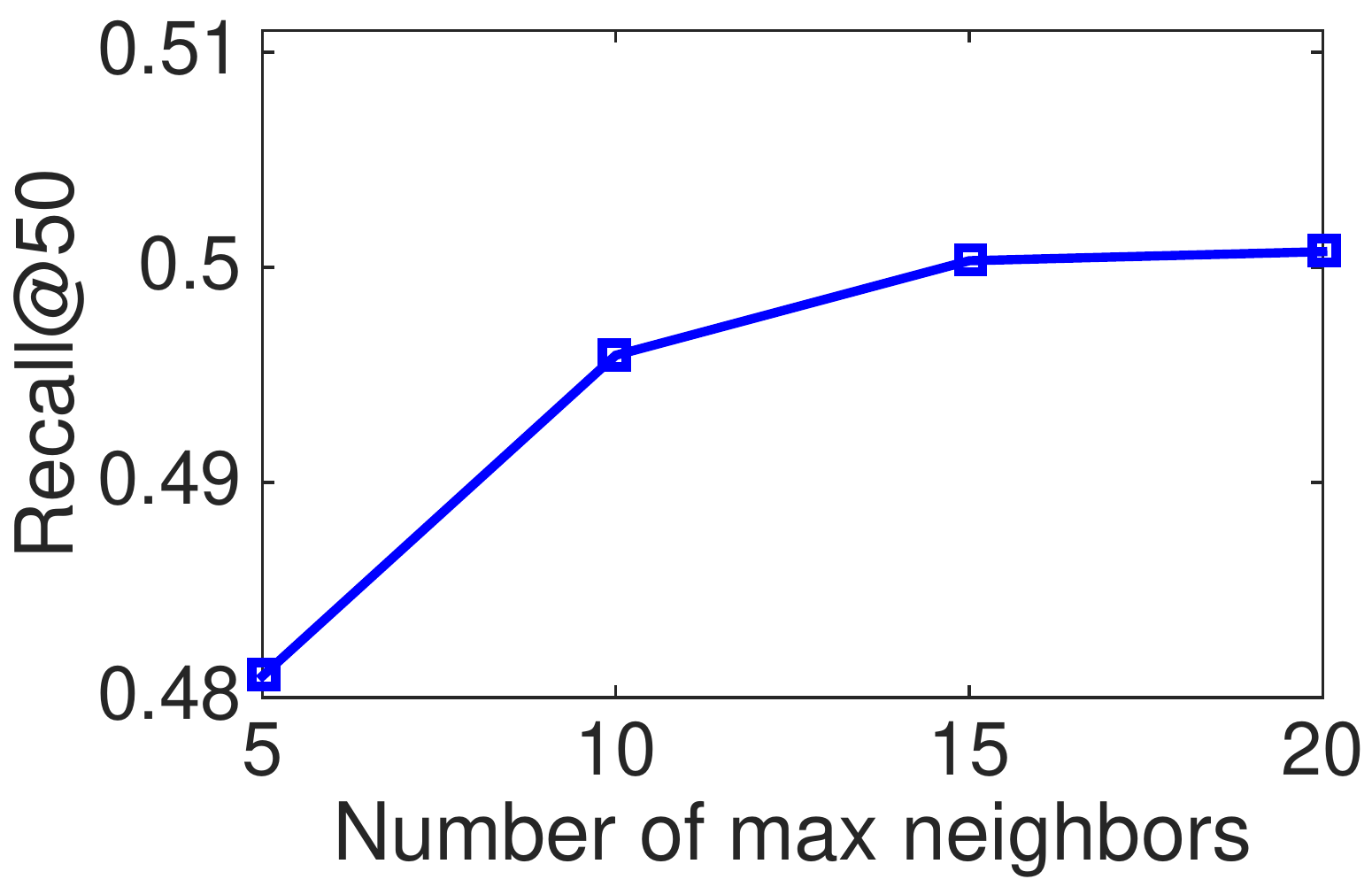}}~~~~
\subfigure[\emph{Lastfm} $K=200$] { \includegraphics[width=3.7cm]{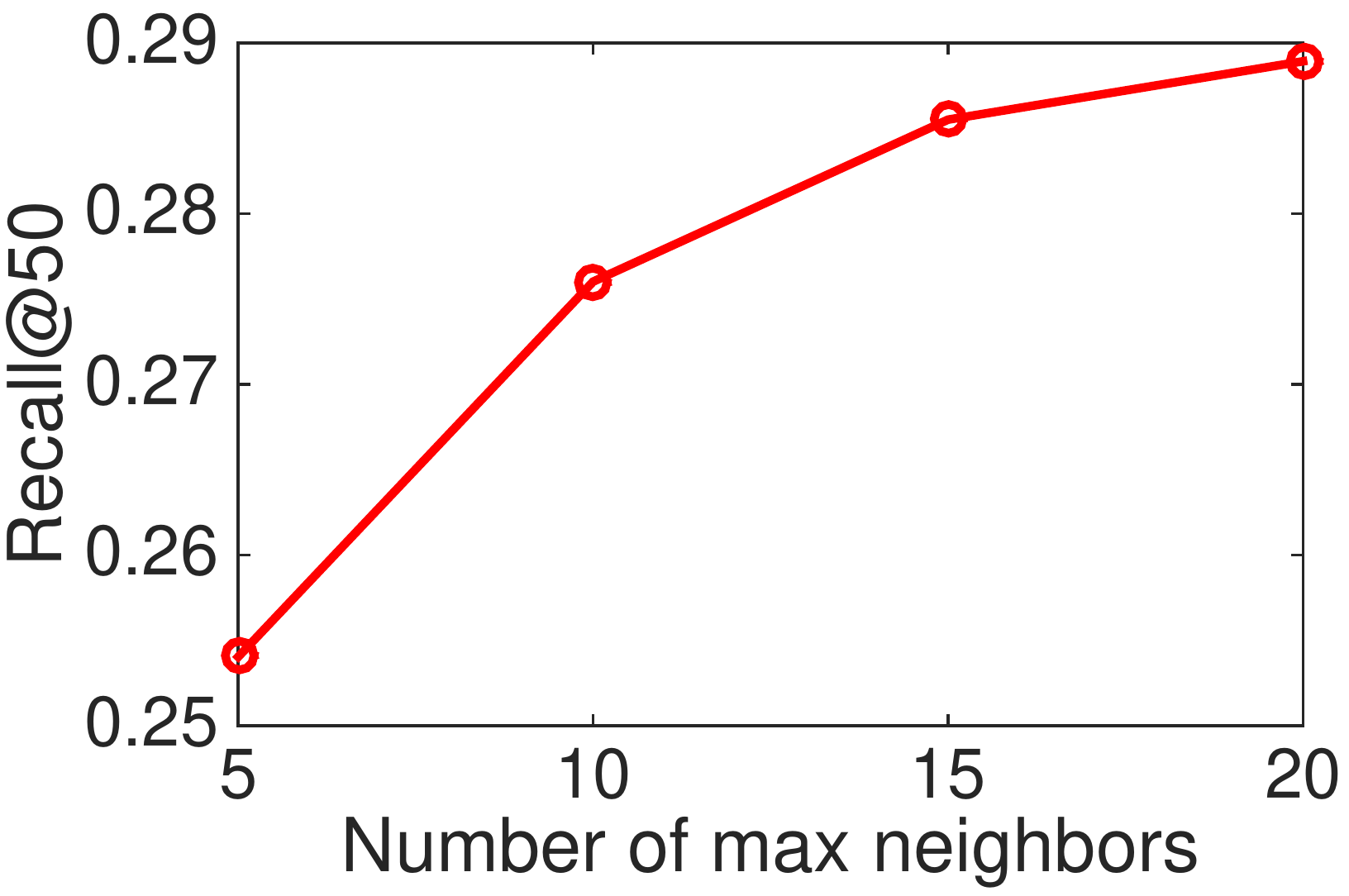}}
%\vskip -0.15in
\caption{Effect of affinigh graph (the number of max neighbors) on RSCGM.}
\label{graphneighbor}
%\vskip -0.2in
\end{figure}

As we described in Section 5.2, the affinigy graph significantly affects the final performance, and therefore, we first study the effect of user/item affinity graphs on our model performance. 
The user/item affinity graphs aim to capture the affinitive relationships between users/items. 
The more precise the built user/item affinity graphs capture all the latent affinitive relationships between users/items, the better our model learns user/item latent factors. 
We change the user/item affinity graphs by varying the number of max neighbors. 
Figure \ref{graphneighbor} shows the effect of the number of max neighbors of the affinity graphs on our model performance, where we set $K=200$. 
From it, we observe that with the increase of the number of max neighbors of the affinity graphs, our model performance first increases and then becomes stable. 
The results indicate that the quality of the affinity graphs are important for our model performance.

\begin{figure}
\centering
\subfigure [\emph{Precision@50}]{ \includegraphics[width=3.7cm]{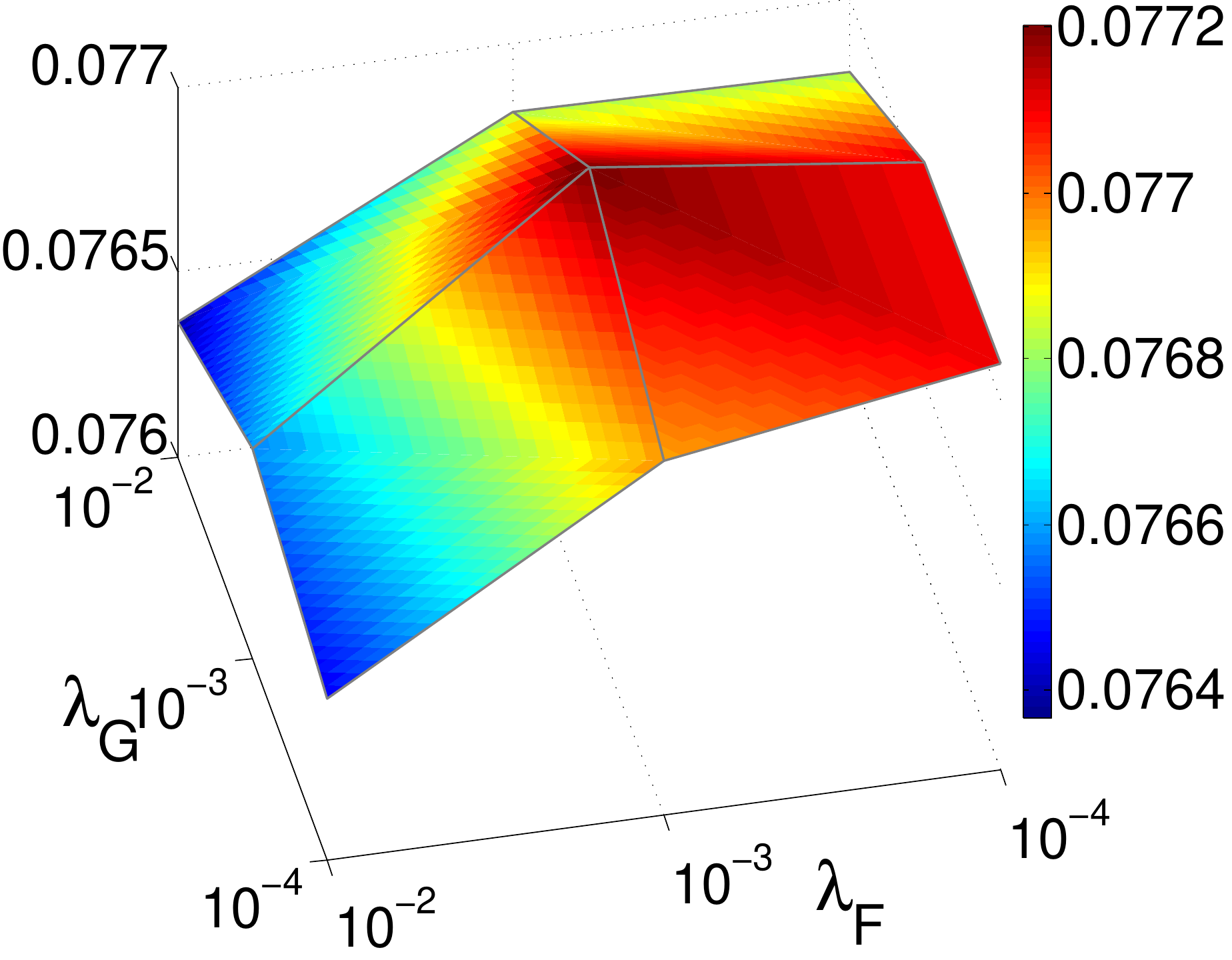}}~~~~
\subfigure[\emph{Recall@50}] { \includegraphics[width=3.7cm]{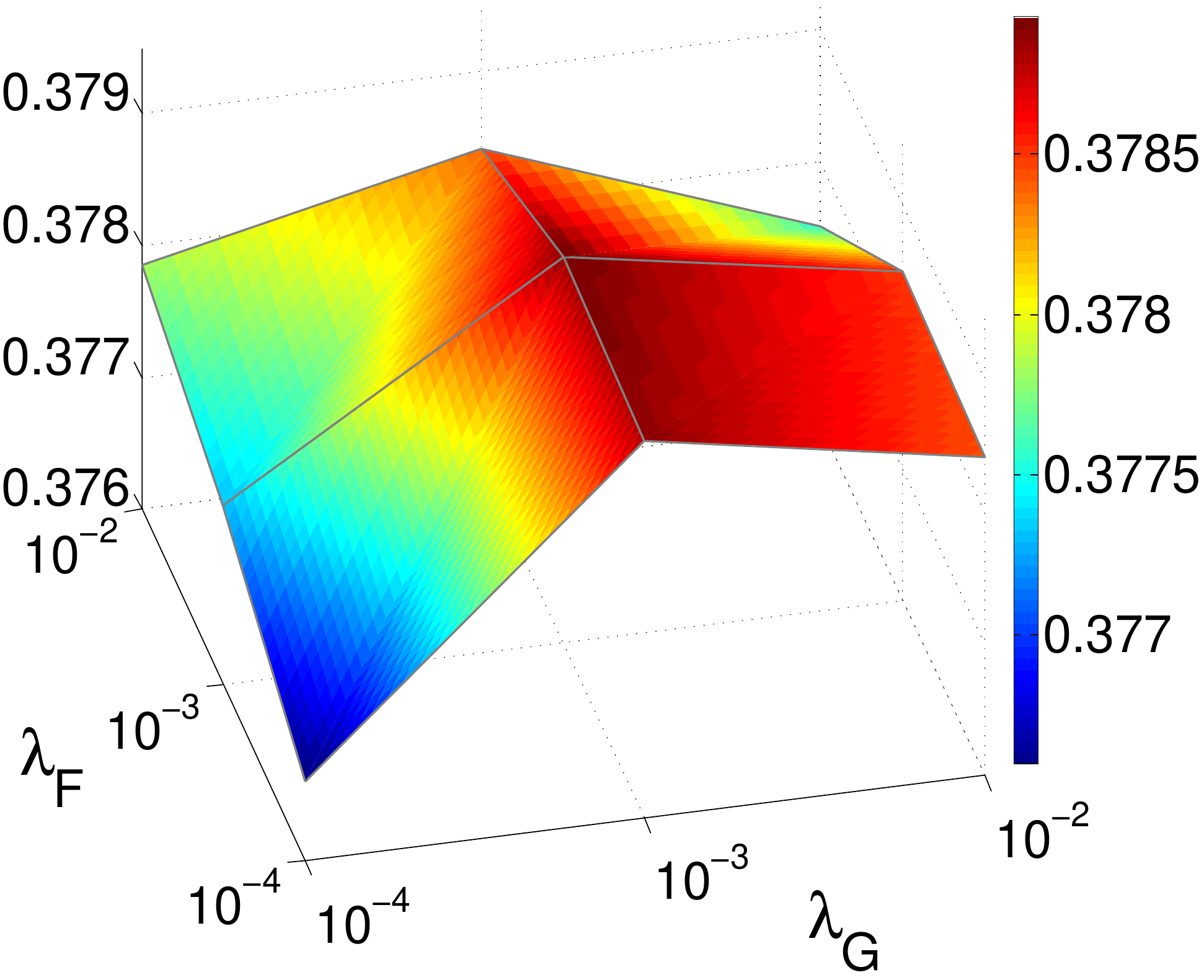}}
%\vskip -0.15in
\caption{Effect of smoothness degree parameters $\lambda_F$ and $\lambda_G$ on RSCGM. Dataset used: \emph{Delicious}.}
\label{graphfg}
%\vskip -0.2in
\end{figure}

\begin{figure}[!htb]
\centering
\subfigure [\emph{Precision@50}]{ \includegraphics[width=3.7cm]{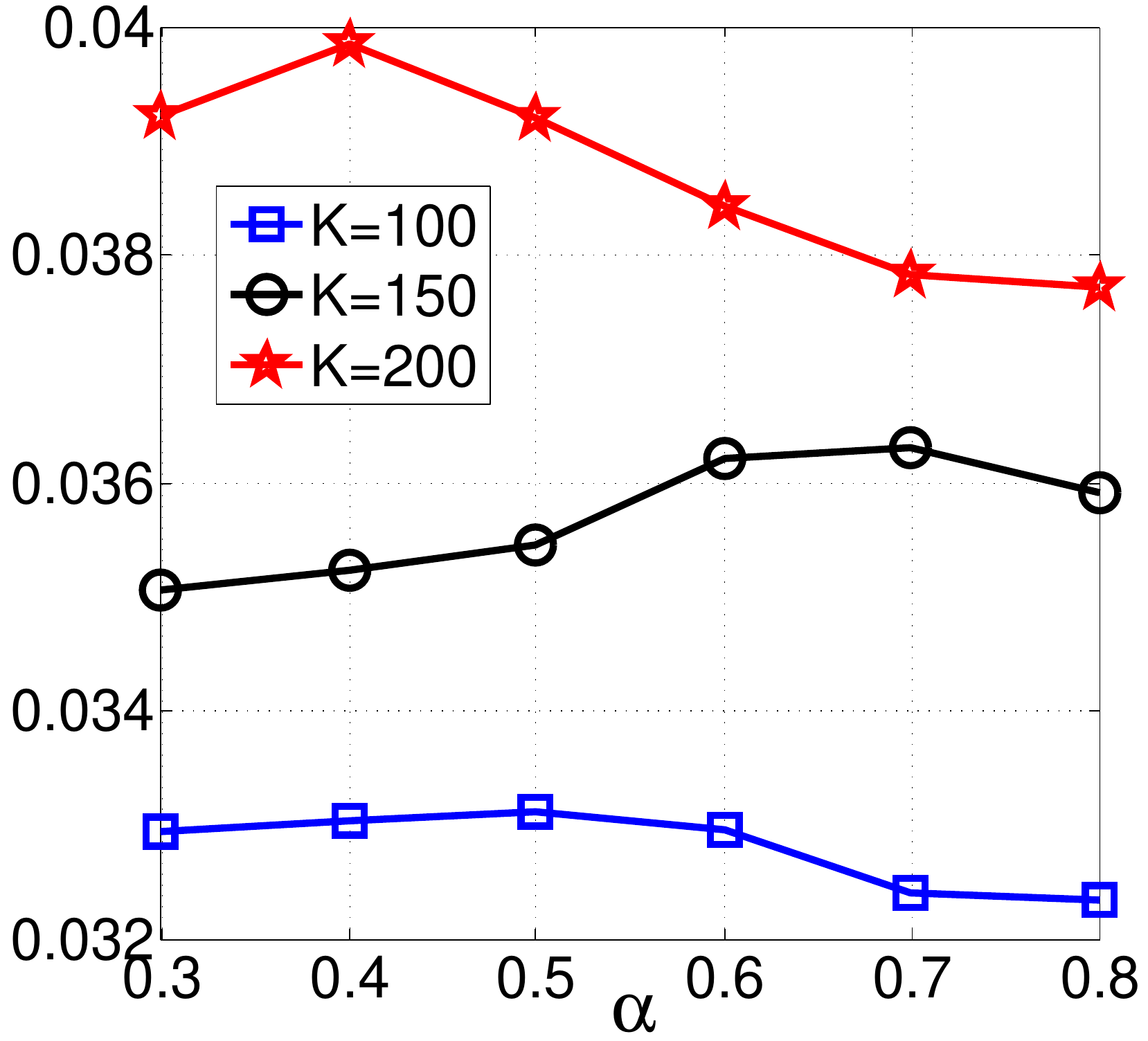}}
\subfigure[\emph{Recall@50}] { \includegraphics[width=3.9cm]{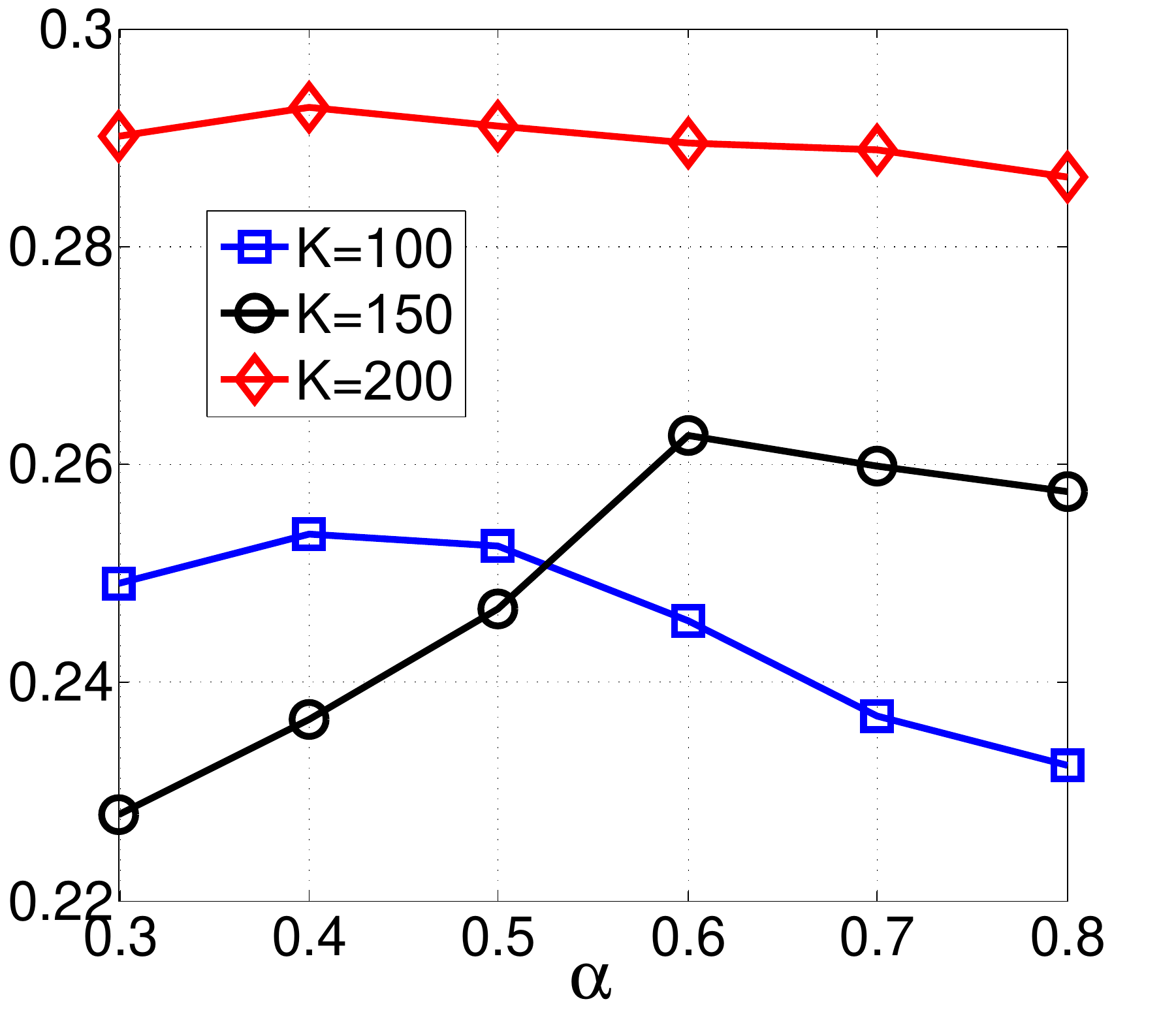}}
%\vskip -0.15in
\caption{Effect of smoothness confidence decay parameters $\alpha$ on RSCGM. Dataset used: \emph{Lastfm}.}
\label{alpha}
%\vskip -0.1in
\end{figure}

We then study the effect of the smoothness degree parameters, i.e., $\lambda_F$ and $\lambda_G$, on our model performance. By doing this, we set $\alpha=1$, $d=0$, which means that we only propagate the known ratings to their direct U-I pairs. 
%Take Figure \ref{propagation} for example, we only propagate the rating of $R_{42}$ to its direct neighbors, i.e., $r_{32}$ and $4_{52}$.
Figure \ref{graphfg} shows the effect of $\lambda_F$ and $\lambda_G$ on RSCGM, where we set $K=100$. We can see that RSCGM achieves the best \emph{Precision} and \emph{Recall} performance when $\lambda_F=\lambda_G=0.001$ on \emph{Delicious}.
The results indicate that smoothness on both user and item affinity graphs contribute to model performance, which proves the effectiveness of our joint-smoothness idea.

We finally study the effect of the smoothness confidence decay parameter, i.e., $\alpha$, on our model performance. By doing this, we set $\lambda_F$ and $\lambda_G$ to the best values obtained above. Figure \ref{alpha} shows the effect of $\alpha$ on RSCGM. From it, we can see that, with the increase of $\alpha$, the performance first increases, and then decreases after a certain threshold. The explanation is: (1) when $\alpha$ is small, the  smoothness confidence $\alpha^{|d|+1}$ will be too small to propagate ratings. That is, known ratings will only be propagated to short path neighbors and this may cause information loss; (2) On the contrary, with a big $\alpha$, known ratings will be propagated to their long path neighbors, and this may cause data noise due to the unreliable affinity graphs.

\subsection{Computation Time}

\begin{figure}[!htb]
\centering
\subfigure[$\alpha=0.3$] { \includegraphics[width=4cm]{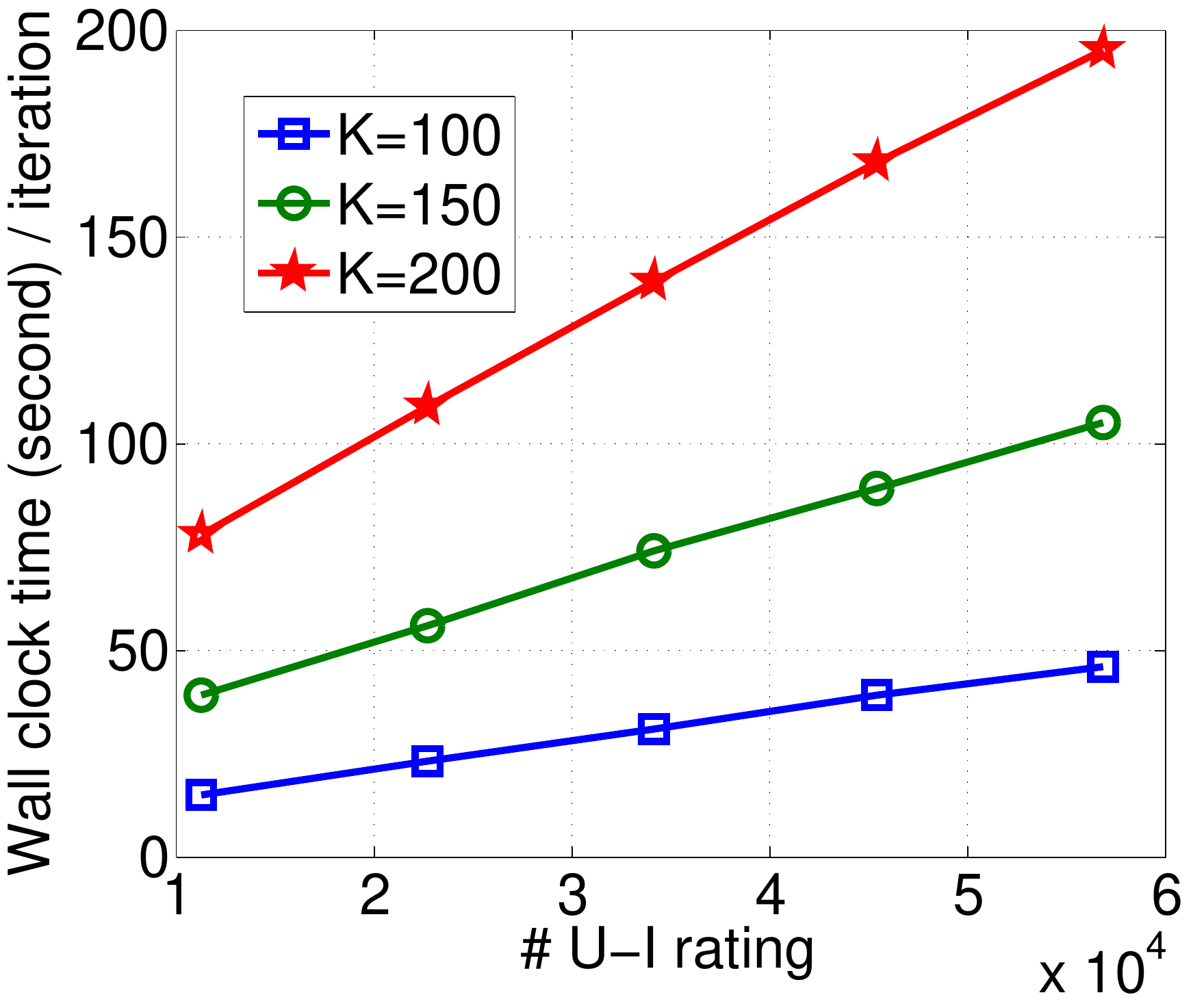}}
\subfigure [\emph{$\alpha=0.4$}]{ \includegraphics[width=4cm]{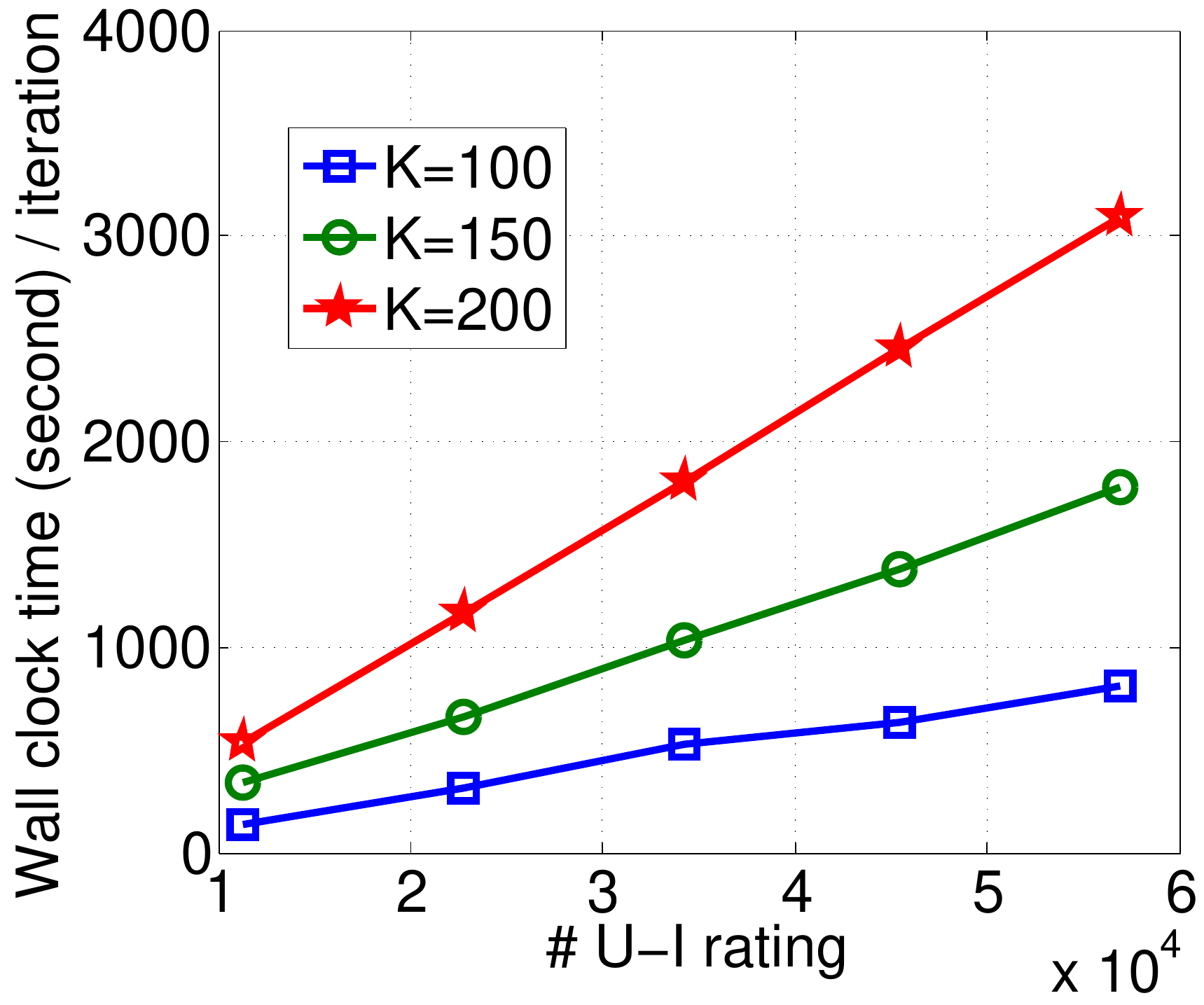}}
%\vskip -0.15in
\caption{Wall-clock time with different size of training data on \emph{Lastfm}.}
\label{time}
%\vskip -0.2in
\end{figure}

We run our model on a PC with 2.33 GHz Intel(R) Xeon(R) CPU and 8Gb RAM.
Figure \ref{time} shows the wall-clock running time per iteration of our proposed model with different size of training data on the \emph{Lastfm}. The same as we analyzed in Section \hyperref[parameter-learning]{\ref{parameter-learning}}, %with the increase of $\alpha$, the runtime decrease, and
the runtime does increase linearly with the observed data size.
Besides, further observation shows that, the bigger $K$ is, the bigger the runtime increase rate is. This is because the runtime can be seen as a linear function of the observed data size, and its slope is $(\sum{\overline{F}} + \sum{\overline{G}}) K$. With a relative stable value of $\sum{\overline{F}} + \sum{\overline{G}}$, the bigger $K$ is, the bigger the increase slope is.

\section{Conclusion}\label{conclusion}

In this paper, we have proposed a probabilistic chain graph models to marry SSL with LFM to improve recommendation performance by alleviating the data sparsity problem.
The proposed CGM is a combination of Bayesian network and Markov random field.
We use the dimensionality reduction idea of LFMs in the Bayesian network, and use the smoothness idea of SSL in Markov random field.
We have proposed to perform joint smoothness instead of pairwise smoothness to save affinity graph build time. 
We also have proposed a confidence-aware approach to realize joint-smoothness to address the affinity unreliable problem in RS.
Our proposed approach realized the ideas of both SSL and LFM, and addresses the challenges of adopting SSL in RS, and thus possesses the merits of both LFM and SSL. 
We have conducted experiments on four popular real world datasets, and the experimental results showed that our approach significantly outperforms the state-of-the-art recommendation approaches, especially in data sparsity scenarios.

\bibliographystyle{ACM-Reference-Format}
\bibliography{reference}

\end{document}